\documentclass[12pt]{article}

\date{}
\begin{document}
\title{A matrix approach for computing extensions of argumentation frameworks}
\author{ {X\small{u} \large{Y}\small{uming}}
\thanks{\footnotesize  Corresponding author. E-mail: xuyuming@sdu.edu.cn }\\
{\small $School \ of \ Mathematics, Shandong \ University, Jinan,  China$}
 }
\maketitle

\hspace{5mm} \begin{minipage}{12.5cm}

\begin{center}{\small \bf Abstract} \end{center}

{\small The matrices and their sub-blocks are introduced into the study of determining  various extensions in the sense of Dung's theory of argumentation frameworks. It is showed that each argumentation framework has its matrix representations, and the core semantics defined by Dung can be characterized by specific sub-blocks of the matrix. Furthermore, the elementary permutations of a matrix are employed by which an efficient matrix approach for finding out all extensions under a given semantics is obtained. Different from several established approaches, such as the graph labelling algorithm, Constraint Satisfaction Problem algorithm, the matrix approach not only put the mathematic idea into the investigation for finding out various extensions, but also completely achieve the goal to compute all the extensions needed. }

\vspace{2mm} \par {\small \it  \hspace{-1mm} Keywords:}
{\small Argumentation framework; extension; matrix; sub-block; permutation}

\end{minipage}\\ \\

\hspace{-10mm} {\bf \large 1. Introduction}

\vspace{5mm}
In recent years, the area of argumentation begins to become increasingly central as a core study within Artificial Intelligence. A number of papers investigated and compared the properties of different semantics which have been proposed for argumentation frameworks (AFs, for short) as introduced by Dung [9, 5, 4, 10, 7]. On the algorithms of abstract argumentation frameworks, Modgil and Caminada[15] list the following questions:

1. For a given semantics, "global" questions concerning the existence and construction of extensions can be addressed:

(a) Does an extension exist?

(b) Given an extension

(c) Given all extensions

2. For a given semantics, "local" questions concerning the existence and construction of extensions, relative to a set $A \in \mathcal{A}$ can be addressed:

(a) Is A contained in an extension?

(b) Is A contained in all extensions?

(c) Is A attacked by an extension?

(d) Is A attacked by all extensions?

(e) Given an extension containing A.

(f) Given all extensions containing A.

(g) Given an extension that attacks A.

(h) Given all extensions that attack A.

\hspace{-8mm} They reviewed two important approaches, graph labelling algorithm and argument game algorithm, which can be used to answer a selection of the above questions with respect to finite argumentation frameworks. Another excellent approach was established by Amgoud and Devred which encodes  argumentation frameworks as the Constraint Satisfaction Problems[1].

\vspace{2mm} Our purpose is to introduce the matrix as a tool into the research of argumentation frameworks. In fact, there is a natural way to assign a matrix of order $n$ for each argumentation framework with $n$ arguments. Each element of the matrix has only two possible values: one and zero, where one represents the attack relation and zero represents the non-attack relation between two arguments (they can be the same one). The matrix can be thought to be a representation of the argumentation framework, since it contains all the information of the argumentation framework. Firstly, we define several sub-blocks of the matrix corresponding to related extensions of an argumentation framework, and give the matrix criterions to determine whether a conflict-free set is a stable extension, admissible extension or complete extension. Then, we employ the elementary permutation of matrices to handle the matrix of an argumentation framework, and obtain a matrix approach for finding out all extensions under a given semantics. It consists of three steps: Defining a family of basic sets, by which we can find out all the conflict-free sets of an argumentation framework; For each conflict-free set, turning the matrix of the argumentation framework into a norm form through a sequence of dual interchanges; Determining what extension the conflict-free set is according to the corresponding criterions of related extensions.

This paper is organized as follows: Section 2 briefly presents Dung's theory on abstract argumentation frameworks. Section 3 introduces the matrix representation of argumentation frameworks, as well as the concept of sub-blocks of a matrix. Section 4 discusses the criterion of determining conflict-free sets, and defines the concept of basic sets by which we can find out all conflict-free sets of an argumentation framework in a programmed way. From Section 5 to Section 7, we address the criterions of determining the core extensions of an argumentation framework, such as stable extension, admissible extension and complete extension. In Section 8, we mainly devote to establish a systematic matrix approach, by which all the questions mentioned by Modgil and Caminada can be solved. Certainly, this approach can be easily realized in practical computing with no technical difficulty. \\

\hspace{-10mm} {\bf \Large 2. Dung's theory of argumentation}

\vspace{5mm} Argumentation is a general approach to model defeasible reasoning and justification in Artificial Intelligence. So far, many theories of argumentation have been established. Among them, Dung's theory of AFs is quite influence. In fact, it is abstract enough to manage without any assumption on the nature of arguments and the attack relation between arguments. Let us first recall some basic notion in Dung's theory of AFs. We restrict them to finite argumentation frameworks.

An argumentation framework is a pair $F = (A, R)$, where $A$ is a finite set of arguments and $R \subset A \times A$ represents the attack-relation. For $S \subset A$, we say that

(1) $S$ is conflict-free in $(A, R)$ if there are no $a, b \in S$ such that $(a, b) \in R$;

(2) $a \in A$ is defeated by $S$ in $(A, R)$ if there is $b \in S$ such that $(b, a) \in R$;

(3) $a \in A$ is defended by $S$ in $(A, R)$ if for each $b \in A$ with $(b, a) \in R$,  we

\hspace{6mm} have $b$ is defeated by $S$ in $(A, R)$.

(4) $a \in A$ is acceptable with respect to $S$ if for each $b \in A$ with $(b, a) \in R$,

\hspace{6mm} there is some $c \in S$ such that $(c, b) \in R$.

The conflict-freeness, as observed by Baroni and Giacomin[2] in their study of evaluative criteria for extension-based semantics, is viewed as a minimal requirement to be satisfied within any computationally sensible notion of "collection of justified arguments". However, it is too weak a condition to be applied as a reasonable guarantor that a set of arguments is "collectively acceptable".

Semantics for AFs can be given by a function $\sigma$ which assigns each argumentation framework $F = (A, R)$ a collection $\mathcal{S} \subset 2^{A}$ of extensions. Here, we mainly focus on the semantic $\sigma \in \{s, a, p, c, g, i, ss, e\}$ for stable, admissible, preferred, complete, grounded, ideal, semi-stable and eager extensions, respectively.

\vspace{2mm} \hspace{-10mm} \textbf{Definition 1}[16] Let $F = (A, R)$ be an argumentation framework and $S \in A$.

(1) $S$ is a stable extension of $F$, $i.e.$, $S \in s(F)$, if $S$ is conflict-free in $F$

\hspace{6mm} and each $a \in A \setminus S$ is defeated by $S$ in $F$.

(2) $S$ is an admissible extension of $F$,  $i.e.$, $S \in a(F)$, if $S$ is conflict-free

\hspace{6mm} in $F$ and each $a \setminus S$ is defended by $S$ in $F$.

(3) $S$ is a preferred extension of $F$,  $i.e.$, $S \in p(F)$, if $S \in a(F)$ and for

\hspace{6mm} each $T \in a(F)$, we have $S \not\subset T$.

(4)  $S$ is a complete extension of $F$,  $i.e.$, $S \in c(F)$, if $S \in a(F)$ and for

\hspace{6mm} each $a \in A$ defended by $S$ in $F$, we have $a \in S$.

(5)  $S$ is a grounded extension of $F$,  $i.e.$, $S \in g(F)$, if $S \in c(F)$ and for

\hspace{6mm} each $T \in c(F)$, we have $T \not\subset S$.

(6)  $S$ is an ideal extension of $F$,  $i.e.$, $S \in i(F)$, if $S \in a(F)$, $S \subset \cap \{T: $

\hspace{6mm} $T \in p(F)\}$ and for each $U \in a(F)$ such that $U \subset \cap \{T: T \in p(F)\}$,

\hspace{6mm} we have $S \not\subset U$.

(7)  $S$ is a semi-stable extension of $F$,  $i.e.$, $S \in ss(F)$, if $S \in a(F)$ and for

\hspace{6mm} each $T \in a(F)$, we have $R^{+}(S) \not\subset R^{+}(T)$, where $R^{+}(U) = U \cap \{b: $

\hspace{6mm} $(a, b) \in R, a \in U\}$.

(8)  $S$ is a eager extension of $F$,  $i.e.$, $S \in e(F)$, if $S \in a(F)$, $S \subset \cap \{T: $

\hspace{6mm} $T \in ss(F)\}$ and for each $U \in a(F)$ such that $U \subset \cap \{T: T \in ss(F)\}$,

\hspace{6mm} we have $S \not\subset U$.

\vspace{2mm} Note that, there are some basic properties for any argumentation framework $F = (A, R)$ and semantic $\sigma$. If $\sigma \in \{a, p, c, g\}$, then we have $\sigma(F) \neq \emptyset$. And if $\sigma \in \{g, i, e\}$, then $\sigma(F)$ contains exactly one extension. Furthermore, the following relations hold for each argumentation framework $F = (A, R)$:

\vspace{2mm}\hspace{40mm}     $s(F) \subseteq p(F) \subseteq c(F) \subseteq a(F)$.

\vspace{2mm} Since every extension of an AF under the standard semantics (stable, preferred, complete and grounded) is an admissible set, the concept of admissible extension plays an important role in the study of AFs. \\

\hspace{-10mm} {\bf \Large 3.  The matrix of an argumentation framework}

\vspace{5mm} We know that the directed graph is a traditional tool in the research of AFs, and has the feature of visualization [8, 11, 12]. It is widely used for modeling and analyzing AFs. In this section, we shall concern another way, that is the matrix representation of AFs. Except for the visualization, the matrix also has the advantage of computability in analyzing the properties of AFs and computing their various extensions.

An $m \times n$ matrix $A$ is a rectangular array of numbers, consisting of $m$ rows and $n$ columns, denoted by

\[
  A = \left(\begin{array}{cccccc}
  a_{1,1}&a_{1,2}&.&.&.&a_{1,n}\\
  a_{2,1}&a_{2,2}&.&.&.&a_{2,n}\\
  .&.&.&.&.&.\\
  a_{m,1}&a_{m,2}&.&.&.&a_{m,n}
  \end{array}\right).
\]
The $m \times n$ numbers $a_{1,1}, a_{1,2}, ..., a_{m,n}$ are the elements of the matrix $A$. We often called $a_{i,j}$ the $(i,j)$th element, and write $A = (a_{i, j})$ for short. It is important to remember that the first suffix of $a_{i,j}$ indicates the row and the second the column of $a_{i,j}$.

An elementary operation on a matrix is an operation of one of the following three types.

(1) The interchange of two rows (or columns).

(2) The multiplication of a row (or column) by a non-zero scalar.

(3) The addition of a multiple of one row (or column) to another row

\hspace{6mm} (column).

A column matrix is an $n \times 1$ matrix, and a row matrix is an $1 \times n$ matrix, denoted by

\[
  \left(\begin{array}{c}  x_{1}\\ x_{2}\\ .\\ .\\ .\\ x_{n}  \end{array}\right), \left(\begin{array}{cccccc}  x_{1}&x_{2}&.&.&.&x_{n}  \end{array}\right)
\]
respectively. Matrices of both these types can be regarded as vectors and referred to respectively as column vectors and row vectors.
Usually, the $i$th row of a matrix $A$ is denoted by $A_{i, *}$, and the $j$th column of the matrix $A$ is denoted by $A_{*, j}$.\\

\hspace{-10mm} \textbf{Definition 2} Let $A = (a_{i, j})$ be an $n \times m$ matrix, $1 \leq i_{1} < i_{2} < ..., < i_{k} \leq n$ and $1 \leq j_{1} < j_{2} < ..., < j_{h} \leq n$. The the matrix

\[
  \left(\begin{array}{cccccc}
  a_{i_{1}, j_{1}}&a_{i_{1}, j_{2}}&.&.&.&a_{i_{1}, j_{h}}\\
  a_{i_{2}, j_{1}}&a_{i_{2}, j_{2}}&.&.&.&a_{i_{2}, j_{h}}\\
  .&.&.&.&.&.\\
  a_{i_{k}, j_{1}}&a_{i_{k}, j_{2}}&.&.&.&a_{i_{k}, j_{h}}
  \end{array}\right),
\]

\hspace{-8mm} is called a $k \times h$ sub-block of the matrix $A$, and denoted by
$M_{i_{1},i_{2},...,i_{k}}^{i_{1},i_{2},...,i_{k}}$.

\vspace{2mm} In particular, the matrix

\[
  \left(\begin{array}{cccccc}
  a_{i_{1}, i_{1}}&a_{i_{1}, i_{2}}&.&.&.&a_{i_{1}, i_{k}}\\
  a_{i_{2}, i_{1}}&a_{i_{2}, i_{2}}&.&.&.&a_{i_{2}, i_{k}}\\
  .&.&.&.&.&.\\
  a_{i_{k}, i_{1}}&a_{i_{k}, i_{2}}&.&.&.&a_{i_{k}, i_{k}}
  \end{array}\right),
\]

\hspace{-8mm} is called a principal sub-block of order $k$ of the matrix $A$. \\

\hspace{-10mm} \textbf{Definition 3} Let $A = (a_{i, j})$ be an $n \times m$ matrix, $1 \leq i_{1} < i_{2} < ..., < i_{k} \leq n$ and $1 \leq j_{1} < j_{2} < ..., < j_{h} \leq n$. If in the $n \times m$ matrix $A = (a_{i, j})$, we delete the rows and columns which make up the sub-block $M^{i_{1},i_{2},...,i_{k}}_{i_{1},i_{2},...,i_{k}}$, then the remaining elements form an $(n-k) \times (m-k)$ matrix

\[
  \left(\begin{array}{cccccc}
  a_{j_{1}, i_{1}}&a_{j_{1}, i_{2}}&.&.&.&a_{j_{1}, i_{h}}\\
  a_{j_{2}, i_{1}}&a_{j_{2}, i_{2}}&.&.&.&a_{j_{2}, i_{h}}\\
  .&.&.&.&.&.\\
  a_{j_{k}, i_{1}}&a_{j_{k}, i_{2}}&.&.&.&a_{j_{k}, i_{h}}
  \end{array}\right),
\]

\hspace{-8mm} We call this matrix the complementary sub-block of $M^{i_{1},i_{2},...,i_{k}}_{i_{1},i_{2},...,i_{k}}$, and denote it by $\overline{M_{i_{1},i_{2},...,i_{k}}^{i_{1},i_{2},...,i_{k}}}$. \\

A matrix divided by horizontal and vertical lines is called a partition matrix. It can be represented by denoting each part array by a single matrix symbol. For example, a $4 \times 3$ matrix $A = (a_{i, j})$ can be partitioned into the following form

\[
  \left(\begin{array}{cc}
  A_{1, 1}&A_{1, 2}\\
  A_{2, 1}&A_{2, 2}
  \end{array}\right),
\]
where 
\[ 
A_{1,1} = \left(\begin{array}{cc}   a_{1, 1}&a_{1, 2}\\    a_{2, 1}&a_{2, 2}    \end{array}\right), 
A_{1,2} = \left(\begin{array}{cc}   a_{1, 3}\\             a_{2, 3}             \end{array}\right),
A_{2,1} = \left(\begin{array}{cc}   a_{3, 1}&a_{3, 2}\\    a_{4, 1}&a_{4, 2}     \end{array}\right),
A_{2,2} = \left(\begin{array}{cc}   a_{3, 3}\\             a_{4, 3}              \end{array}\right).
\]

Let $F = (A, R)$ be an argumentation framework where $A$ is a finite set. It is obvious that the notation $A = \{a, b, ...\}$ is not convenience when the cardinality of $A$ is too large, so we prefer to denote $A$ by $\{1, 2, ..., n\}$, subsequently.

For the underlying finite set $A$ of $F = (A, R)$, there is no ordering in nature. But, an ordered set can benefit us a lot in many cases. Based on this consideration, we introduce the concept of permutation into our discussion.

A permutation of a finite set $A$ of $n$ elements is a mapping of the set onto itself. The usual method of presenting a permutation is to write down the elements of $A$ in a row in natural order and, under each of them to write down its image. For convenience, we usually write down only the elements of the image in a row without changing their original order. For example, the following is a permutation of the set $A = \{1, 2, 3, 4, 5 \}$

\[
  \left(\begin{array}{ccccc}
  1&2&3&4&5\\
  3&5&2&1&4
  \end{array}\right) = \left(\begin{array}{ccccc}
  3&5&2&1&4
  \end{array}\right).
\]

For any the argumentation framework $F = (A, R)$ with $A = \{1, 2, ..., n\}$, let $(i_{1}, i_{2}, ..., i_{n})$ be a fixed permutation of $A$, then  $F = (A, R)$ can be represented by a Boolean matrix under some simple rules.

\vspace{2mm}
\hspace{-10mm} \textbf{Definition 4} Let $F = (A, R)$ be an argumentation framework with $A = \{1, 2, ..., n\}$. The matrix of $F$ corresponding to the permutation $(i_{1}, i_{2}, ..., i_{n})$ of $A$, denoted by $M(i_{1}, i_{2}, ..., i_{n})$, is a Boolean matrix of order $n$, its elements are determined by the following rules:

(1) $a_{s, t } = 1$ iff $(i_{s}, i_{t}) \in R$;

(2) $a_{s, t} = 0$ iff $(i_{s}, i_{t}) \notin R$.

\vspace{2mm}
\hspace{-10mm} \textbf{Remark} In matrix $M(i_{1}, i_{2}, ..., i_{n})$, the elements of $r$-th row reflect the attack relations of the argument $i_{r}$ to the other arguments, while the elements of $r$-th column reflect the attacked relations of the other arguments to the argument $i_{r}$. So, the argumentation framework $F = (A, R)$ with $A = \{1, 2, ..., n\}$ has many different matrix representations, which depend on the different selection of permutations of $A$.

Although the argumentation framework $F = (A, R)$ has many different matrix representations, all these matrices have the same role in representing  $F = (A, R)$. In other words, they are equivalent in representing $F = (A, R)$. We usually use the matrix $M(1, 2, ..., n)$ corresponding to the natural permutation $(1, 2, ..., n)$ to represent the argumentation framework $F$, and denote it by $M(F)$.

\vspace{2mm}
\hspace{-10mm} \textbf{Example 5} Considering the argumentation framework $F = (A, R)$, where $A = \{1,2,3\}$ and $R = \{(1, 2), (2, 3), (3, 1)\}$. By Definition 4, the matrix of $F$ corresponding to the natural permutation $(i_{1}, i_{2}, i_{3}) = (1, 2, 3)$ is

\[
  M(F) = M(1, 2, 3) = \left(\begin{array}{ccc}
  0&1&0\\
  0&0&1\\
  1&0&0
  \end{array}\right)
\]

\hspace{-8mm} The matrix of $F$ corresponding to the permutation $(i_{1}, i_{2}, i_{3}) = (2, 1, 3)$ is

\[
  M(2, 1, 3) = \left(\begin{array}{ccc}
  0&0&1\\
  1&0&0\\
  0&1&0
  \end{array}\right)
\]

In comparison with directed graph way and logical analysis way, the matrix $M(F)$ of an argumentation framework $F$ has many excellent features. For example, it possess a concise mathematical format and contains all the information of $F$ by combining the arguments with attack relations in a specific manner. Also, we can import the knowledge and method of matrix theory to the research of AFs, or even discover some new knowledge of matrices for the research of AFs. The more important is that a powerful soft for dealing with the computing of matrices has been developed, by which a lot of work can be saved. \\

\hspace{-10mm} {\bf \Large 4.  Determination of the conflict-free sets}

\vspace{5mm} As we have known, there is no efficient method for us to decide a conflict set in an argumentation framework, even we can draw up the directed graph of it. After introducing the matrix of the AF, the situation will be changed completely. Let us see an example, firstly.

\vspace{2mm}
\hspace{-10mm} \textbf{Example 6} Given an argumentation framework $F = (A, R)$, where $A = \{1,2,3,4,5\}$ and $R = \{(1, 2), (2, 3), (2, 5), (4, 3), (5, 4)\}$. Then, we can easily to show that the family of conflict-free sets of $F$ is

     $\{ \emptyset, \{1\}, \{2\}, \{3\}, \{4\}, \{5\}£¬ \{1,3\}, \{1,4\}, \{1,5\}, \{2,4\}, \{3,5\}, \{1,3,5\} \}$,

\hspace{-8mm} by the traditional method of directed graph.

Now, we consider the matrix of $F = (A, R)$ and study its structure from the level of sub-blocks. First, we write out the matrix of $F$ corresponding to the natural permutation $(1,2,3,4,,5)$:

\[
  M(F) = \left(\begin{array}{ccccc}
  0&1&0&0&0\\
  0&0&1&0&1\\
  0&0&0&0&0\\
  0&0&1&0&0\\
  0&0&0&1&0
  \end{array}\right).
\]

\hspace{-8mm} Secondly, we concentrate our attention on the principal sub-blocks of the above matrix. There are five zero principal sub-blocks of order 1:
\[
  M^{1}_{1} = \left(\begin{array}{c}   0    \end{array} \right), M^{2}_{2} = \left(\begin{array}{c}  0  \end{array}  \right), M^{3}_{3} = \left(\begin{array}{c}   0  \end{array}  \right), M^{4}_{4} = \left(\begin{array}{c}   0  \end{array}  \right), M^{5}_{5} = \left(\begin{array}{c}   0  \end{array}  \right),
\]
which correspond to the conflict-free sets $\{1\}$, $\{2\}$, $\{3\}$, $\{4\}$, $\{5\}$, respectively.
There are five zero principal sub-blocks of order 2:
\[
  M^{1,3}_{1,3} = \left(\begin{array}{cc}   0&0\\    0&0    \end{array} \right), M^{1,4}_{1,4} = \left(\begin{array}{cc}   0&0\\    0&0    \end{array}  \right), M^{1,5}_{1,5} = \left(\begin{array}{cc}   0&0\\    0&0    \end{array}  \right), M^{2,4}_{2,4} = \left(\begin{array}{cc}   0&0\\    0&0    \end{array}  \right),
\]

\[
  \hspace{-100mm} M^{3,5}_{3,5} = \left(\begin{array}{cc} 0&0\\ 0&0 \end{array}  \right).
\]
They exactly match with the conflict-free sets $\{1,3\}$, $\{1,4\}$, $\{1,5\}$, $\{2,4\}$, $\{3,5\}$, respectively. Also, there is one zero principal sub-block of order 3:

\[
  M^{1,3,5}_{1,3,5} = \left(\begin{array}{ccc}   0&0&0\\    0&0&0\\   0&0&0    \end{array} \right),
\]
which is followed by the conflict-free sets $\{1, 3, 5\}$.

\vspace{2mm} Note that, the above sub-blocks are all principal sub-blocks which are zero in the matrix $M(F)$, and there is a one to one correspond between the family of all zero principal sub-blocks of $M(F)$ and the family of all conflict-free sets of $F = (A, R)$. In fact, for any argumentation framework $F$ there exists such corresponding relation between the family of all zero principal sub-blocks of $M(F)$ and the family of all conflict-free sets of $F = (A, R)$.

\vspace{2mm}
\hspace{-10mm} \textbf{Definition 7} Let $F = (A, R)$ be an argumentation framework with $A = \{1, 2, ..., n\}$, and $S = \{i_{1}, i_{2}, ..., i_{k}\} \subset A$ satisfying $1 \leq i_{1} < i_{2} < ... < i_{k} \leq n$. The principal sub-block

\[
  M^{i_{1}, i_{2}, ..., i_{k}}_{i_{1}, i_{2}, ..., i_{k}} = \left(\begin{array}{cccccc}
  a_{i_{1}, i_{1}}&a_{i_{1}, i_{2}}&.&.&.&a_{i_{1}, i_{k}}\\
  a_{i_{2}, i_{1}}&a_{i_{2}, i_{2}}&.&.&.&a_{i_{2}, i_{k}}\\
  .&.&.&.&.&.\\
  a_{i_{k}, i_{1}}&a_{i_{k}, i_{2}}&.&.&.&a_{i_{k}, i_{k}}
  \end{array}\right)
\]
of order $k$ in the matrix $M(F)$ is called the $cf$-sub-block of $S$, and denoted by $M^{cf}$ for short.

\vspace{2mm}
\hspace{-10mm} \textbf{Theorem 8} Given an argumentation framework $F = (A, R)$ with $A = \{1, 2, ..., n\}$, then $S = \{i_{1}, i_{2}, ..., i_{k}\} \subset A (1 \leq i_{1} < i_{2} < ... < i_{k} \leq n)$ is a conflict-free set in $F$ iff the $cf$-sub-block $M^{i_{1}, i_{2}, ..., i_{k}}_{i_{1}, i_{2}, ..., i_{k}}$ of $S$ is zero.

\vspace{2mm}
\hspace{-10mm} \textbf{Proof} Assume that $M^{i_{1}, i_{2}, ..., i_{k}}_{i_{1}, i_{2}, ..., i_{k}} = 0$, then for arbitrary $1 \leq s, t \leq k$ we have $a_{i_{s}, i_{t}} = 0$, $i.e.$, $(i_{s}, i_{t}) \notin R$. Thus, $S = \{i_{1}, i_{2}, ..., i_{k}\}$ is a conflict-free set in $F$.

 Suppose $S = \{i_{1}, i_{2}, ..., i_{k}\} \subset A$ is a conflict-free set in $F$, then for arbitrary $1 \leq s, t \leq k$ we have that $(i_{s}, i_{t}) \notin R$, $i.e.$, $a_{i_{s}, i_{t}} = 0$. Therefore, we have $M^{i_{1}, i_{2}, ..., i_{k}}_{i_{1}, i_{2}, ..., i_{k}} = 0$.

\vspace{2mm}
Next, we shall develop a way to find out all the zero principal sub-blocks in the matrix $M(F)$ of an argumentation framework $F$, which is corresponding to all the conflict-free sets of $F$.

\vspace{2mm}
Given an argumentation framework $F = (A, R)$ with $A = \{1, 2, ..., n\}$, let $\mathcal{S}(r)$ denote the family of all conflict-free sets in $F$ whose cardinality are $r (1 \leq r \leq n)$. Then, $\mathcal{S}(1) = \{ \{1\}, \{2\}, ..., \{n\}\}$. Surely, $\mathcal{S}(2)$ can be easily decided according to the matrix $M(F)$ of $F = (A, R)$. In fact, for conflict-free sets which posses more than one element we can find out them all by the following theorem.

\vspace{2mm}
\hspace{-10mm} \textbf{Theorem 9} For any argument $i \in A$ and subset $S \in \mathcal{S}(r)(1 \leq r \leq n)$, let $C(i) = \{j \in A: a_{i, j} = 0 \ or \ a_{j, i} = 0\}$. If $i \notin S$ and $S \subset C(i)$, then $S \cup \{i\}$ is a conflict-free set whose cardinality is $r + 1$.

\vspace{2mm}
\hspace{-10mm} \textbf{Proof} We need only to prove that $a_{i, k} = 0$ and $a_{k, i} = 0$ for each $k \in S$. It is a direct result of the fact $S \subset C(i)$.

\vspace{5mm}
\hspace{-10mm} \textbf{Remark} By Theorem 9, we can find out all the families $\mathcal{S}(1), \mathcal{S}(2), ..., \mathcal{S}(n)$ in the following way: First, determining the sets $C(1), C(2), ..., C(n)$, which will be called the basic sets of the argumentation framework $F = (A, R)$. Secondly, writing out the family $\mathcal{S}(1) = \{ \{1\}, \{2\}, ..., \{n\} \}$ from the values of $a_{i, i}(1 \leq i \leq n)$. Then, by comparing the elements of $\mathcal{S}(1)$ with the basic sets $C(j)(1 \leq j \leq n)$, we can find out all the elements of the family $\mathcal{S}(2)$. The family $\mathcal{S}(3)$ can be decided by comparing the elements of $\mathcal{S}(2)$ with the basic sets $C(j)(1 \leq j \leq n)$ in a similar process, and so on.

\vspace{2mm}
\hspace{-10mm} \textbf{Example 10} Consider the argumentation framework $F = (A, R)$ in Example 8. The matrix of $F$ corresponding to the natural permutation $(1,2,3,4,5)$ is

\[
  M(F) = \left(\begin{array}{ccccc}
  0&1&0&0&0\\
  0&0&1&0&1\\
  0&0&0&0&0\\
  0&0&1&0&0\\
  0&0&0&1&0
  \end{array}\right).
\]

It is obviously that $\mathcal{S}(0) = \{\emptyset \}$ and $\mathcal{S}(1) = \{ \{1\}, \{2\}, ..., \{5\} \}$. Next, we shall computer the family $\mathcal{S}(2)$, $\mathcal{S}(3)$, $\mathcal{S}(4)$ and $\mathcal{S}(5)$.

 First, we list the basic sets $C(1) = \{ 3, 4, 5 \}$, $C(2) = \{ 4 \}$, $C(3) = \{ 1, 5 \}$, $C(4) = \{ 1, 2 \}$ and $C(5) = \{ 1, 3 \}$.

 Second, by the fact that $\{3\}, \{4\}, \{5\} \subset C(1)$, we have $ \{1, 3\}, \{1, 4\},\{1, 5\} \in \mathcal{S}(2)$. Similarly, we can conclude that $ \{2, 4\}, \{3, 5\} \in \mathcal{S}(2)$ by the fact that $ \{ 4 \} \subset C(2)$, and $ \{ 5 \} \subset C(3)$. Note that, $\{ 1 \} \subset C(3)$ will result in that $\{1, 3\} \in \mathcal{S}(2)$, but it is not essential because of the previous case. Similar situation are also arise for $\{ 1\}, \{ 2 \} \subset C(4)$ and $\{ 1\}, \{ 3 \} \subset C(5)$. Therefore, $\mathcal{S}(2) = \{\{1, 3\}, \{1, 4\},\{1, 5\}, \{2, 4\}, \{3, 5\}\}$.

For the family $\mathcal{S}(3)$, we only need to compare $C(1) = \{ 3, 4, 5 \}$ with the elements of $\mathcal{S}(2)$. Since $\{3, 5\} \in \mathcal{S}(2)$ and $\{3, 5\} \subset C(1)$, we have $ \{1, 3, 5\} \in \mathcal{S}(3)$. In fact, we conclude that $\mathcal{S}(3) = \{\{1, 3, 5\} \}$.

Since there are no basic sets containing more than three elements, we claim that $\mathcal{S}(4)$ and $\mathcal{S}(5)$ are both empty sets.

\vspace{5mm}
\hspace{-10mm} \textbf{Remark} In the process to computing the family $\mathcal{S}(i+1)$ from $\mathcal{S}(i)$ and  $C(j)(1 \leq j \leq n)$, we need only to check the relation between the element $S \in \mathcal{S}(i)$ and the basic set $C(j)$ satisfying $j < max \{k: k \in S\}$. In fact, if $S \in \mathcal{S}(i)$ with $j \notin S$ and there is some $k \in S$ such that $k < j$, then $S \subset C(j)$ will implies $S \cup \{j\} \in \mathcal{S}(i+1)$. But this result can also be obtained when we compare the element $(S \cup \{j\}) \setminus \{k\}$ of $\mathcal{S}(i)$ and the basic set $C(k)$. Of course, $(S \cup \{j\}) \setminus \{k\} \in \mathcal{S}(i)$ and $(S \cup \{j\}) \setminus \{k\} \subset C(k)$ comes from the fact $S \cup \{j\} \in \mathcal{S}(i+1)$.  \\

\hspace{-10mm} {\bf \Large 5.  Determination of the stable extensions}

\vspace{5mm}
\hspace{-10mm} \textbf{Example 11} We continuous to study the argumentation framework $F = (A, R)$, where $A = \{1,2,3,4,5\}$ and $R = \{(1, 2), (2, 3), (2, 5), (4, 3), (5, 4)\}$ in Example 10. Since the stable extension is firstly a conflict-free set, we can look for the stable extension from the collection

\vspace{2mm}
$\{\emptyset, \{1\}, \{2\}, \{3\}, \{4\}, \{5\}£¬\{1,3\}, \{1,4\}, \{1,5\}, \{2,4\}, \{3,5\}, \{1,3,5\} \}$

\vspace{2mm}
\hspace{-8mm} of conflict-free sets. In fact, the set $S = \{1,3,5\}$ is the only stable extension in $F = (A, R)$ by a short discussion.

\vspace{2mm}
Let us turn our attention to the matrix

\[
  M(F) = \left(\begin{array}{ccccc}    0&1&0&0&0\\      0&0&1&0&1\\     0&0&0&0&0\\     0&0&1&0&0\\     0&0&0&1&0     \end{array}\right).
\]
of the $F = (A, R)$, and try to find the information contained in $M(F)$ which insure the conflict-free set $S = \{1,3,5\}$ is a stable extension.

\vspace{2mm}\hspace{-8mm} Note that, $(1,2) \in R$ implies the argument $2$ is defeated by $S = \{1,3,5\}$, and $(5,4) \in R$ implies the argument $4$ is defeated by $S = \{1,3,5\}$. It is exactly the two results which make the conflict-free set $S$ to be a stable extension. By Definition 5, the conditions $(1,2) \in R$ and $(5,4) \in R$ are represented in the form of $a_{1,2} = 1$ and $a_{5,4} = 1$ in the matrix $M(F)$, respectively. And, the argument $2$ is defeated by $S = \{1,3,5\}$ is equivalent to the column vector

\[
  \left(\begin{array}{ccc}    a_{1,2}\\      a_{3,2}\\     a_{5,2}     \end{array}  \right )\neq 0,
\]
the argument $4$ is defeated by $S = \{1,3,5\}$ is equivalent to the column vector

\[
  \left(\begin{array}{ccc}    a_{1,4}\\      a_{3,4}\\     a_{5,4}     \end{array}  \right )\neq 0.
\]

It is not difficult to see that $(1,2) \in R$ and $(5,4) \in R$ is not sufficient for the conflict-free set $S$ to be stable. The sufficient and necessary conditions for the conflict-free set $S$ to be a stable extension are that, the argument $2$ is defeated by $S = \{1,3,5\}$ and the argument $4$ is defeated by $S = \{1,3,5\}$. These facts are reflected in the matrix $M(F)$ as follows:

(1) $a_{1,2} = 1$ and $a_{5,4} = 1$ is not the sufficient condition for the conflict-free set $S$ to be stable,

(2) The sufficient and necessary conditions for the conflict-free set $S$ to be a stable extension are

\[
  \left(\begin{array}{ccc}    a_{1,2}\\      a_{3,2}\\     a_{5,2}     \end{array}  \right )\neq 0, \ \ \ and \ \ \ \left(\begin{array}{ccc}    a_{1,2}\\      a_{3,2}\\     a_{5,2}     \end{array}  \right )\neq 0.
\]

Considering the sub-block

\[
  M^{s} = \left(\begin{array}{ccc}    a_{1,1}&a_{1,3}&a_{1,5}\\      a_{3,1}&a_{3,3}&a_{3,5}\\
   a_{5,1}&a_{5,3}&a_{5,5}     \end{array}  \right ) = \left(\begin{array}{ccc}    0&0&0\\      0&0&0\\
   0&0&0     \end{array}  \right ),
\]
which insures of the conflict-freeness of the set $S = \{1,3,5\}$, if we combine the above two column vectors into the following matrix

\[
  \left(\begin{array}{ccc}    a_{1,2}&a_{1,4}\\      a_{3,2}&a_{3,4}\\
   a_{5,2}&a_{5,4}     \end{array}  \right ),
\]
then it is also a sub-block of $M(F)$ and responds to the determining of the stableness of $S = \{1,3,5\}$.

\vspace{2mm} The above analysis motivates us to propose the following definition, and provide a matrix way to determine whether a conflict-free set is a stable extension of an argumentation framework.

\vspace{2mm}
\hspace{-10mm} \textbf{Definition 12} Let $F = (A, R)$ be an argumentation framework with $A = \{1, 2, ..., n\}$, and $S = \{i_{1}, i_{2}, ..., i_{k}\} \subset A (1 \leq i_{1} < i_{2} < ... < i_{k} \leq n)$ a stable extension of $F$. The $k \times h$ sub-block

\[
  M^{i_{1}, i_{2}, ..., i_{k}}_{j_{1}, j_{2}, ..., j_{h}} = \left(\begin{array}{cccccc}
  a_{i_{1}, j_{1}}&a_{i_{1}, j_{2}}&.&.&.&a_{i_{1}, j_{h}}\\
  a_{i_{2}, j_{1}}&a_{i_{2}, j_{2}}&.&.&.&a_{i_{2}, j_{h}}\\
  .&.&.&.&.&.\\
  a_{i_{k}, j_{1}}&a_{i_{k}, j_{2}}&.&.&.&a_{i_{k}, j_{h}}
  \end{array}\right)
\]
in the matrix $M(F)$ is called the $s$-sub-block of $S$ and denoted by $M^{s}$ for short, where $\{j_{1}, j_{2}, ..., j_{h}\} = A \setminus S (1 \leq j_{1} < j_{2} < ... < j_{h} \leq n)$.

In other words, the elements at the intersections of rows $i_{1}, i_{2}, ..., i_{k}$ and columns $j_{1}, j_{2}, ..., j_{h}$ in the matrix $M(F)$ form the $s$-sub-block $M^{i_{1}, i_{2}, ..., i_{k}}_{j_{1}, j_{2}, ..., j_{h}}$ of the set $S$.

\vspace{2mm}
\hspace{-10mm} \textbf{Theorem 13} Given an argumentation framework $F = (A, R)$ with $A = \{1, 2, ..., n\}$, then conflict-free set $S = \{i_{1}, i_{2}, ..., i_{k}\}  \subset A (1 \leq i_{1} < i_{2} < ... < i_{k} \leq n)$ is a stable extension in $F$ iff each column vector of the $s$-sub-block $M^{s}$ of $S$ is non-zero, where $A \setminus S = \{j_{1}, j_{2}, ..., j_{h}\}$ and $1 \leq j_{1} < j_{2} < ... < j_{h} \leq n$.

\vspace{2mm}
\hspace{-10mm} \textbf{Proof} Let $S$ be a conflict-free set and $A \setminus S = \{j_{1}, j_{2}, ..., j_{h}\}$, then we need only to prove that every element of $A \setminus S (1 \leq t \leq h)$ is defeated by $S$ in $F$ iff all column vectors of the $s$-sub-block $M^{s}$ of $S$ are non-zero.

Assume that every element of $A \setminus S (1 \leq t \leq h)$ is defeated by $S$ in $F$. Take any column vector $M_{*, j_{t}} (1 \leq t \leq h)$ of the $s$-sub-block $M^{s} = M^{i_{1}, i_{2}, ..., i_{k}}_{j_{1}, j_{2}, ..., j_{h}}$ of $S$, then $j_{t} \in A \setminus S$. By the assumption, there is some element $i_{r} \in S (1 \leq r \leq k)$ which attacks the argument $j_{t}$. It follows that $(i_{r}, j_{t}) \in R$. This is reflcted by $a_{i_{r}, j_{t}} = 1$ in the matrix $M(F)$, and thus the column vector $M^{s}_{*, j_{t}}$ of the $s$-sub-block $M^{s} = M^{i_{1}, i_{2}, ..., i_{k}}_{j_{1}, j_{2}, ..., j_{h}}$ of $S$ is non-zero.

Conversely, suppose that each column vector of the $s$-sub-block $M^{s}$ of $S$ is non-zero. Take any element $j_{t} \in A \setminus S (1 \leq t \leq h)$, then $M^{s}_{*, j_{t}}$ is a column vector of the $s$-sub-block $M^{s} = M^{i_{1}, i_{2}, ..., i_{k}}_{j_{1}, j_{2}, ..., j_{h}}$ of $S$. By the hypothesis, we know that $M_{*, j_{t}}$ is non-zero. Therefore, there is some $i_{r} \in S (1 \leq r \leq k)$ such that $a_{i_{r}, j_{t}} = 1$, i.e., $(i_{r}, j_{t}) \in R$. This means that the argument $i_{r}$ attacks the argument $j_{t}$ of $S$ in $F$, and thus we claim that $j_{t}$ is defeated by $S$ in $F$. \\

\hspace{-10mm} {\bf \Large 6.  Determination of the admissible extensions}

\vspace{5mm}
\hspace{-10mm} \textbf{Example 14} Let us return to the argumentation framework $F = (A, R)$, where $A = \{1,2,3,4,5\}$ and $R = \{(1, 2), (2, 3), (2, 5), (4, 1), (4, 3), (5, 4)\}$ in Example 10. Since an admissible extension is necessarily a conflict-free set, we can look for the admissible extension from the collection

$\{\emptyset, \{1\}, \{2\}, \{3\}, \{4\}, \{5\}£¬\{1,3\}, \{1,5\}, \{2,4\}, \{3,5\}, \{1,3,5\} \}$

\hspace{-8mm} of conflict-free sets. By definition, it is easy to check that $\{1,5\}$ and $\{1,3,5\}$ are the all admissible extensions in $F = (A, R)$.

In order to find the matrix way to determine the admissible extensions, we mainly focus our attention on the admissible extension $S = \{1, 5\}$ which is not stable.

Note that, $(4, 1) \in R$, $(3, 1) \in R$, $(5, 3) \in R$ and $(5, 4) \in R$ implies  that the argument $1$ is defended by $S = \{1, 5\}$, and $(2, 5) \in R$ and $(1, 2) \in R$ implies  that the argument $5$ is defended by $S = \{1, 5\}$. It follows that $S = \{1, 5\}$ is an admissible extensions in $F$. Interestingly, we have anther explanation for $S = \{1, 5\}$ to be admissible. That is the attacker $2$ of $S$ is defeated by $S$, the attacker $3$ of $S$ is defeated by $S$, and the attacker $4$ of $S$ is defeated by $S$.

Now, let us turn our attention to the matrix

\[
  M(F) = \left(\begin{array}{ccccc}    0&1&0&0&0\\      0&0&1&0&1\\     1&0&0&0&0\\     1&0&1&0&0\\     0&0&1&1&0     \end{array}\right).
\]
of $F$, and try to find out the corresponding representation of the above conditions in the matrix $M(F)$.

Firstly, $(3, 1) \in R$ and $(5, 3) \in R$ are reflected by $a_{3, 1} = 1$ and $a_{5, 3} = 1$ in the matrix $M(F)$ respectively. $(2, 5) \in R$ and $(1, 2) \in R$ are reflected by $a_{2, 5} = 1$ and $a_{1, 2} = 1$ in the matrix $M(F)$ respectively. While, $(4, 1) \in R$ and $(5, 4) \in R$ are reflected by $a_{4, 1} = 1$ and $a_{5, 4} = 1$ in the matrix $M(F)$ respectively.

Secondly, $a_{2, 5} = 1$, $a_{3, 1} = 1$ and $a_{4, 1} = 1$ imply the row vectors $(a_{2,1}, a_{2,5})$, $(a_{3,1}, a_{3,5})$ and $(a_{4,1}, a_{4,5})$ are all non-zero. Moreover, $a_{1, 2} = 1$, $a_{5, 3} = 1$ and $a_{5, 4} = 1$ imply the following column vectors are all non-zero:

\[
  \left(\begin{array}{ccc}    a_{1,2}\\      a_{5,2}   \end{array}  \right ), \left(\begin{array}{ccc}    a_{1,3}\\      a_{5,3}   \end{array}  \right ), \left(\begin{array}{ccc}    a_{1,4}\\      a_{5,4}   \end{array}  \right ).
\]

Thirdly, the attacker $2$ of $S$ is defeated by $S$ is equivalent to that when the row vectors $(a_{2,1}, a_{2,5}) \neq 0$ then the column vector

\[
  \left(\begin{array}{ccc}    a_{1,2}\\      a_{5,2}   \end{array}  \right ) \neq 0.
\]
The attacker $3$ of $S$ is defeated by $S$ is equivalent to that when the row vectors $(a_{3,1}, a_{3,5}) \neq 0$ then the column vector

\[
  \left(\begin{array}{ccc}    a_{1,3}\\      a_{5,3}   \end{array}  \right ) \neq 0,
\]
The attacker $4$ of $S$ is defeated by $S$ is equivalent to that when the row vectors $(a_{4,1}, a_{4,5}) \neq 0$ then the column vector

\[
  \left(\begin{array}{ccc}    a_{1,4}\\      a_{5,4}   \end{array}  \right ) \neq 0.
\]

Finally, if we combine the row vectors $(a_{2,1}, a_{2,5})$, $(a_{3,1}, a_{3,5})$ and $(a_{4,1}, a_{4,5})$, then the matrix

\[
  \left(\begin{array}{ccc}    a_{2,1}&a_{2,5}\\   a_{3,1}&a_{3,5}\\  a_{4,1}&a_{4,5}  \end{array}  \right )
\]
is a sub-block of $M(F)$. While, if we combine the column vectors

\[
  \left(\begin{array}{ccc}    a_{1,2}\\      a_{5,2}   \end{array}  \right ), \left(\begin{array}{ccc}    a_{1,3}\\      a_{5,3}   \end{array}  \right ) \ \ \ and \ \ \  \left(\begin{array}{ccc}    a_{1,4}\\      a_{5,4}   \end{array}  \right ),
\]
then the matrix

\[
  \left(\begin{array}{ccc}    a_{1,2}&a_{1,3}&a_{1,4}\\    a_{5,2}&a_{3,3}&a_{5,4}  \end{array}  \right )
\]
is a sub-block of $M(F)$.

\vspace{2mm} The above analysis motivates us to propose the following definition, and provide a matrix way to determine whether a conflict-free set is an admissible extensions of an argumentation framework.

\vspace{2mm}
\hspace{-10mm} \textbf{Definition 15} Let $F = (A, R)$ be an argumentation framework with $A = \{1, 2, ..., n\}$, and $S = \{i_{1}, i_{2}, ..., i_{k}\} \subset A (1 \leq i_{1} < i_{2} < ... < i_{k} \leq n)$ is an admissible extension of $F$. The $h \times k$ sub-block

\[
  M^{j_{1}, j_{2}, ..., j_{h}}_{i_{1}, i_{2}, ..., i_{k}} = \left(\begin{array}{cccccc}
  a_{j_{1}, i_{1}}&a_{j_{1}, i_{2}}&.&.&.&a_{j_{1}, i_{k}}\\
  a_{j_{2}, i_{1}}&a_{j_{2}, i_{2}}&.&.&.&a_{j_{2}, i_{k}}\\
  .&.&.&.&.&.\\
  a_{j_{h}, i_{1}}&a_{j_{h}, i_{2}}&.&.&.&a_{j_{h}, i_{k}}
  \end{array}\right)
\]
of the matrix $M(F)$ is called the $a$-sub-block of $S$ and denoted by $M^{a}$, where $\{j_{1}, j_{2}, ..., j_{h}\} = A \setminus S$ and $1 \leq j_{1} < j_{2} < ... < j_{h} \leq n$.

In other words, the elements at the intersection of rows $j_{1}, j_{2}, ..., j_{h}$ and columns $i_{1}, i_{2}, ..., i_{k}$ in the matrix $M(F)$ form the $a$-sub-block $M^{j_{1}, j_{2}, ..., j_{h}}_{i_{1}, i_{2}, ..., i_{k}}$ of $S$.

Note that, there is a natural relation between the $a$-sub-block $M^{j_{1}, j_{2}, ..., j_{h}}_{i_{1}, i_{2}, ..., i_{k}}$ and the $s$-sub-block $M^{i_{1}, i_{2}, ..., i_{k}}_{j_{1}, j_{2}, ..., j_{h}}$ in matrix theory. Namely, the $a$-sub-block $M^{j_{1}, j_{2}, ..., j_{h}}_{i_{1}, i_{2}, ..., i_{k}}$ of $S$ is precisely the complementary sub-block of the $s$-sub-block $M^{i_{1}, i_{2}, ..., i_{k}}_{j_{1}, j_{2}, ..., j_{h}}$  of $S$ in the matrix $M(F)$.

\vspace{2mm}
\hspace{-10mm} \textbf{Theorem 16} Given an argumentation framework $F = (A, R)$ with $A = \{1, 2, ..., n\}$, then a conflict-free set $S = \{i_{1}, i_{2}, ..., i_{k}\} \subset A (1 \leq i_{1} < i_{2} < ... < i_{k} \leq n$ is an admissible extension iff the column vector of the $s$-sub-block $M^{s}$ of $S$ corresponding to
the non-zero row vector of the $a$-sub-block $M^{a}$ of $S$ is non-zero, where $A \setminus S = \{j_{1}, j_{2}, ..., j_{h}\}$ and $1 \leq j_{1} < j_{2} < ... < j_{h} \leq n$ .

\vspace{2mm}
\hspace{-10mm} \textbf{Proof} Let $S$ be a conflict-free set and $A \setminus S = \{j_{1}, j_{2}, ..., j_{h}\}$. We need only to prove that every $i_{r} \in S (1 \leq r \leq k)$ is defended by $S$ iff the column vector of $s$-sub-block $M^{s}$ of $S$ corresponding to the non-zero row vector of the $a$-sub-block $M^{a}$ of $S$ is non-zero

Assume that every $i_{r} \in S (1 \leq r \leq k)$ is defended by $S$. If the row vector $M^{a}_{t, *} (1 \leq t \leq h)$ of the $a$-sub-block $M^{a} = M^{j_{1}, j_{2}, ..., j_{h}}_{i_{1}, i_{2}, ..., i_{k}}$ of $S$ is non-zero, then there is some $i_{r} (1 \leq r \leq k)$ such that $a_{j_{t}, i_{r}} =1$. It follows that $(j_{t}, i_{r}) \in R$, $i.e.$, the argument $i_{r}$ is attacked by the argument $j_{t}$. By the assumption, there is some $i_{q} \in S (1 \leq q \leq k)$ which attacks the argument $j_{t}$, i.e., $(i_{q}, j_{t}) \in R$. This is reflected by $a_{i_{q}, j_{t}} = 1$ in the matrix $M(F)$. Obviously, $a_{i_{q}, j_{t}}$ is an element of the column vector $M^{s}_{*, t}$ of $M^{s}$. Therefore, the column vector $M^{s}_{*, t}$ of the $s$-sub-block $M^{i_{1}, i_{2}, ..., i_{k}}_{j_{1}, j_{2}, ..., j_{h}} = M^{s}$ of $S$ is non-zero.

Conversely, suppose any column vector of the $s$-sub-block $M^{s}$ of $S$ corresponding to the non-zero row vector of the $a$-sub-block $M^{s}$ of $S$ is non-zero. Let $i_{r} \in S (1 \leq r \leq k)$, which is attacked by some $j_{t} \in A \setminus S (1 \leq t \leq h)$. Then, $(j_{t}, i_{r}) \in R$, which is reflected by $a_{j_{t}, i_{r}} = 1$ in the matrix $M(F)$. It follows that the row vector $M^{a}_{t, *}$ of the $a$-sub-block $M^{a} = M^{j_{1}, j_{2}, ..., j_{h}}_{i_{1}, i_{2}, ..., i_{k}}$ of $S$ is non-zero. By the assumption, the corresponding column vector $M^{s}_{*, t}$ of the $s$-sub-block $M^{s} = M^{i_{1}, i_{2}, ..., i_{k}}_{j_{1}, j_{2}, ..., j_{h}}$ of $S$ is non-zero. Therefore, there is some $i_{q} \in S (1 \leq q \leq k)$ such that $a_{i_{q}, j_{t}} = 1$ in the matrix $M(F)$. Correspondingly, we have that $(i_{q}, j_{t}) \in R$, and thus the argument $j_{t}$ is attacked by the argument $i_{q} \in S$. To sum up, the argument $i_{r} \in S$ is defended by $S$ in $F$. \\

\vspace{2mm}
\hspace{-10mm} \textbf{Remark:} The fact that any stable extension must be admissible is clearly expressed by the properties of $s$-sub-blocks in the matrix $M(F)$. In other words, the condition every column vector of the $s$-sub-block $M^{s}$ of $S$ are non-zero is stronger than that the column vector of the $s$-sub-block $M^{s}$ of $S$ corresponding to the non-zero row vector of the $a$-sub-block $M^{a}$ of $S$ is non-zero. \\

\hspace{-10mm} {\bf \Large 7.  Determination of the complete extensions}

\vspace{5mm}
\hspace{-10mm} \textbf{Example 17} Consider the argumentation framework $F = (A, R)$, where $A = \{1,2,3,4,5\}$ and $R = \{(1, 4), (2, 1), \{2,3\}, (2, 4), (2, 5), (3, 2), (4,1)\}$. Since the admissible extension is necessarily a conflict-free set, we can find out all admissible extensions from the collection of conflict-free sets

\hspace{-8mm} $\{\emptyset, \{1\}, \{2\}, \{3\}, \{4\}, \{5\}, \{1,3\}, \{1,5\}, \{3,4\}, \{3,5\}, \{4,5\}, \{1,3,5\},  \{3,4,5\} \}$.

\hspace{-8mm} By the directed graph of $F$, it is easy to check that $\{1,3\}$, $\{3,4\}$, $\{3,5\}$, $\{1,3,5\}$ and $\{3,4,5\}$ are all the stable extensions of $F$. Furthermore, one can verify that $\{3,5\}$ is the only complete extension which are not stable, $\{2\}$, $\{3\}$, $\{4\}$ are all the admissible extensions which are not complete.

\vspace{2mm} In order to confirm that the admissible extension $S = \{3,5\}$ is complete, we need to give the reasons which support that the arguments $1, 2, 4$ not defended by $S = \{3,5\}$. There are two cases for us to deal with. One is for the argument $2$. It is attacked by $3$ because of $(3, 2) \in R$, and thus is not defended by $S = \{3,5\}$. Another is for the arguments $1$ and $4$ which has no attacker in $S$. The argument $1$ has two attackers $2$ and $4$ in light of $(2,1), (4, 1) \in R$. Since $(3,2) \in R$, the attacker $2$ of the argument $1$ is attacked by $3$. But, $(3,4), (5,4) \notin R$ implies that the attacker $4$ of the argument $1$ is not attacked by any elements $S = \{3,5\}$. Therefore, the argument $1$ is not defended by $S$. Similar analysis indicates that the argument $4$ is also not defended by $S$.

Note that, in the above discussion the fact that the attacker $2$ of the argument $1$ is attacked by $3$ is not the key for which the argument $1$ is not defended by $S$. So, we can omit the handling of this situation.

\vspace{2mm} Next, we will analysis the expressions of $\{3,5\}$ in the matrix $M(F)$ of $F$ (as a complete extension but not stable), and extract the matrix way to decide that an admissible extension is complete. Let us firstly write out the matrix of the argumentation framework $F = (A, R)$:

\[
  M(F) = \left(\begin{array}{ccccc}    0&0&0&1&0\\      1&0&1&1&1\\     0&1&0&0&0\\     1&0&0&0&0\\     0&0&0&0&0     \end{array}\right).
\]

For the argument $2$ of the first case above, $(3, 2) \in R$ is represented in the form $a_{3,2} = 1$, which results in that the column vector

\[
  \left(\begin{array}{ccc}    a_{3,2}\\      a_{5,2}   \end{array}  \right ) \neq 0.
\]
For the argument $1$ of the second case above, we first note that the column vector

\[
  \left(\begin{array}{ccc}    a_{3,1}\\    a_{5,1}  \end{array}  \right ) = 0.
\]
Furthermore, $(4, 1) \in R$ is represented in the form $a_{4, 1} = 1$, which results in that the column vector

\[
  \left(\begin{array}{ccc}    a_{1,1}\\    a_{2,1}\\      a_{4,1}   \end{array}  \right ) \neq 0.
\]
And, $(3, 4), (5, 4) \notin R$ is represented in the form $a_{3, 4} = 0, a_{5, 4} = 0$, which results in that the column vectors

\[
  \left(\begin{array}{ccc}    a_{3,4}\\    a_{5,4}  \end{array}  \right ) = 0.
\]

\vspace{2mm} For the argument $4$ of the second case above, we can find out its representation in the matrix $M(F)$ in a similar way.

Similar as the $s$-sub-block and the $a$-sub-block, if we combine the column vectors

\[
  \left(\begin{array}{ccc}    a_{1,1}\\    a_{2,1}\\      a_{4,1}   \end{array}  \right ), \left(\begin{array}{ccc}    a_{1,2}\\    a_{2,2}\\      a_{4,2}   \end{array}  \right ) \ \ \ and \ \ \  \left(\begin{array}{ccc}    a_{1,4}\\    a_{2,4}\\      a_{4,4}   \end{array}  \right ),
\]
then we obtain a sub-block

\[
  \left(\begin{array}{ccc}    a_{1,1}&a_{1,2}&a_{1,4}\\      a_{2,1}&a_{2,2}&a_{2,4}\\    a_{4,1}&a_{4,2}&a_{4,4}     \end{array}\right).
\]
which is the key to determine the completeness of $S$.

To sum up, $S_{1} = \{3, 5\}$ is a complete extension can be verified by the following facts contained in the matrix $M(F)$:

(1) When the column vectors of the sub-block

\[
  M^{s} = \left(\begin{array}{ccc}    a_{3,1}&a_{3,2}&a_{3,4}\\      a_{5,1}&a_{5,2}&a_{5,4}     \end{array}\right)
\]
are zero, the corresponding column vectors of the sub-block

\[
  \left(\begin{array}{ccc}    a_{1,1}&a_{1,2}&a_{1,4}\\      a_{2,1}&a_{2,2}&a_{2,4}\\    a_{4,1}&a_{4,2}&a_{4,4}     \end{array}\right)
\]
are non-zero.

(2) The column $4$ of $M^{s}$ corresponding to the row number where $a_{4,1} = 1$ is at is zero, and the column $1$ of $M^{s}$ corresponding to the row number where $a_{1,4} = 1$ is at is zero.

\vspace{2mm} This motivation makes us to propose the following definition, and provide a matrix way to determine whether a conflict-free set is a complete extension of an argumentation framework.

\vspace{2mm}
\hspace{-10mm} \textbf{Definition 18} Let $F = (A, R)$ be an argumentation framework with $A = \{1, 2, ..., n\}$, and $S = \{i_{1}, i_{2}, ..., i_{k}\} \subset A (1 \leq i_{1} < i_{2} < ... < i_{k} \leq n)$ is a complete extension of $F$. The sub-block

\[
  M^{j_{1}, j_{2}, ..., j_{h}}_{j_{1}, j_{2}, ..., j_{h}} = \left(\begin{array}{cccccc}
  a_{j_{1}, i_{1}}&a_{j_{1}, i_{2}}&.&.&.&a_{j_{1}, i_{k}}\\
  a_{j_{2}, i_{1}}&a_{j_{2}, i_{2}}&.&.&.&a_{j_{2}, i_{k}}\\
  .&.&.&.&.&.\\
  a_{j_{h}, i_{1}}&a_{j_{h}, i_{2}}&.&.&.&a_{j_{h}, i_{k}}
  \end{array}\right)
\]
of order $h$ in the matrix of $M(F)$ is called the $c$-sub-block of $S$ and denoted by $M^{c}$ for short, where $\{j_{1}, j_{2}, ..., j_{h}\} = A \setminus S$ and $1 \leq j_{1} < j_{2} < ... < j_{h} \leq n$.

In other words, the elements appearing at the intersection of rows $j_{1}, j_{2}, ..., j_{h}$ and the same number of columns in the matrix $M(F)$ form the $c$-sub-block $M^{j_{1}, j_{2}, ..., j_{h}}_{j_{1}, j_{2}, ..., j_{h}}$ of $S$.

Note that, the $c$-sub-block $M^{c} = M^{j_{1}, j_{2}, ..., j_{h}}_{j_{1}, j_{2}, ..., j_{h}}$ of $S$ is exactly the complementary sub-block of the $s$-sub-block $M^{s} = M^{i_{1}, i_{2}, ..., i_{k}}_{i_{1}, i_{2}, ..., i_{k}}$ of $S$, in the matrix $M(F)$ of $F = (A, R)$.

\vspace{2mm}
\hspace{-10mm} \textbf{Lemma 19} Let $F = (A, R)$ be an argumentation framework with $A = \{1, 2, ..., n\}$, then $S = \{i_{1}, i_{2}, ..., i_{k}\} \subset A (1 \leq i_{1} < i_{2} < ... < i_{k} \leq n)$ is a complete extension of $F$ iff $S$ is an admissible extension and each argument $j_{t} \in S (1 \leq t \leq h)$ is not defended by $S$ in $F$.

\vspace{2mm}
\hspace{-10mm} \textbf{Theorem 20} Given an argumentation framework $F = (A, R)$ with $A = \{1, 2, ..., n\}$, the admissible extension $S = \{i_{1}, i_{2}, ..., i_{k}\} \subset A (1 \leq i_{1} < i_{2} < ... < i_{k} \leq n)$ is complete in $F$ iff each column vector of $c$-sub-block $M^{c}$ of $S$ corresponding to the zero column vector of the $s$-sub-block $M^{s}$ of $S$ is non-zero, and this column vector has at least one non-zero element such that the column vector of the $s$-sub-block $M^{s}$ of $S$ corresponding to the row index of it is zero, where $\{j_{1}, j_{2}, ..., j_{h}\} = A \setminus S$ and $1 \leq j_{1} < j_{2} < ... < j_{h} \leq n$.

\vspace{2mm}
\hspace{-10mm} \textbf{Proof} Let $S$ be an admissible extension and $A \setminus S = \{j_{1}, j_{2}, ..., j_{h}\}$, we need only to prove that every $j_{t} \in S (1 \leq t \leq h)$ is not defended by $S$ in $F$ iff the condition in the theorem is hold.

Necessity. Let the column vector $M^{s}_{*, t}$ of the $s$-sub-block $M^{i_{1}, i_{2}, ..., i_{h}}_{j_{1}, j_{2}, ..., j_{h}}$ of $S$ is zero, where $1 \leq t \leq h$. If the corresponding column vector $M^{c}_{*, t}$ in the $c$-sub-block $M^{j_{1}, j_{2}, ..., j_{k}}_{j_{1}, j_{2}, ..., j_{h}}$ of $S$ is zero, then $a_{i, j_{t}} = 0$ for each $1 \leq i \leq n$. This implies that $(i, j_{t}) \notin R$, i.e., the argument $j_{t}$ is dot attacked by $i$, for each $1 \leq i \leq n$. So, there is no argument of $A$ which attacks $j_{t}$, and thus the argument $j_{t} \in A \setminus S$ is defended by $S$, a contradiction with the completeness of $S$. Therefore, each column vector of $c$-sub-block $M^{c}$ of $S$ corresponding to the zero column vector of the $s$-sub-block $M^{s}$ of $S$ is non-zero.

Suppose the column vector $M^{c}_{*, t}$ of the $c$-sub-block $M^{c}$ of $S$ corresponding to the zero column vector $M^{s}_{*, t}$ of the $s$-sub-block $M^{s}$ of $S$ is non-zero where $1 \leq t \leq h$, but every column vector of the $s$-sub-block $M^{s}$ of $S$ corresponding to the row index of the non-zero element of $M^{c}_{*, t}$ is non-zero, we claim that the argument $j_{t} \in A \setminus S$ is defended by $S$, which contradicts with the completeness of $S$.  In fact, $M^{s}_{*, t}$ indicates that $a_{i_{r}, j_{t}} = 0$, i.e., the argument $i_{r}$ does not attack $j_{t}$ for each $1 \leq r \leq k$. So, the attackers of $j_{t}$ must be in the set $A \setminus S$. Let the argument $j_{v}(1 \leq v \leq h)$ be an attacker of $j_{t}$, then $(j_{v}, j_{t}) \in R$, i.e., $a_{j_{v}, j_{t}} = 1$. By the assumption, the column vector $M^{s}_{*, v}$ of the $s$-sub-block $M^{s}$ of $S$ is non-zero. This implies that there is some $i_{r}(1 \leq r \leq k)$ such that $a_{i_{r}, j_{v}} = 1$, i.e., $(i_{r}, j_{v}) \in R$. Thus, the attacker $j_{v}$ of the argument $j_{t}$ is attacked by an element $i_{r}$ of the set $S$.

Sufficiency. Let $j_{t} \in A \setminus S(1 \leq t \leq h)$, we should prove that $j_{t}$ is not defended by $S$. If there is some argument $i_{r} \in S(1 \leq r \leq k)$ which attacks $j_{t}$, then we have done. Otherwise, $(i_{r}, j_{t}) \in R$, i.e., $a_{i_{r}, j_{t}} = 1$ for each $1 \leq r \leq k$, and thus the column vector $M^{s}_{*, t}$ of $s$-sub-block $M_{s} = M^{i_{1}, i_{2}, ..., i_{h}}_{j_{1}, j_{2}, ..., j_{h}}$ of $S$ is zero. According to the assumption, the column vector $M^{c}_{*, t}$ of the $c$-sub-block $M^{c} = M^{j_{1}, j_{2}, ..., j_{h}}_{j_{1}, j_{2}, ..., j_{h}}$ of $S$ is non-zero, and it has one non-zero element, say $a_{j_{l}, j_{t}} = 1$, such that the column vector $M^{s}_{*, l}$ of the $s$-sub-block $M^{s}$ of $S$ is zero. This indicates that $(j_{l}, j_{t}) \in R$ and $(i_{r}, j_{l}) \notin R$ for each $1 \leq r \leq k$. Thus, the attacker $j_{l}$ of the argument $j_{t}$ does not be attacked by any element of $S$. Therefore, the argument $j_{t}$ is not defended by $S$.   \\

\hspace{-10mm} {\bf \Large 8. A matrix approach for computing extensions}

\vspace{4mm} In matrix theory, the interchange between different two rows (or columns) is one kind of the elementary operations on a matrix, by which a matrix can be reduce to a simply form. In this section, corresponding to the conflict-free set $S$ found out in Section 4 we shall first turn the matrix $M(F)$ of an argumentation framework $F$ into a norm form which is efficient enough for us to determine whether the set $S$ is a stable (admissible, complete) extension, by a sequence of interchanges between different two rows (or columns). Then, combining with the results obtained in the above Sections we put foreword a matrix approach for computing all extensions under a given semantics.

\vspace{2mm} Given an argumentation framework $F = (A, R)$ with $A = \{1, 2, ..., n\}$, let $M(i_{1}, ..., i_{k}, ..., i_{l}, ..., i_{n})(1 < ... < k < ... < l < ... < n)$ be the matrix of $F$ corresponding to the permutation $(i_{1}, ..., i_{k}, ..., i_{l}, ..., i_{n})$. A dual interchange of the matrix $M(i_{1}, ..., i_{k}, ..., i_{l}, ..., i_{n})$ between $k$ and $l$, denoted by $k \rightleftharpoons l$, consists of two interchanges: interchanging row $k$ and row $l$; interchanging column $k$ and column $l$.

\vspace{2mm}
\hspace{-10mm} \textbf{Lemma 21} Given an argumentation framework $F = (A, R)$ with $A = \{1, 2, ..., n\}$. Let $M(i_{1}, ..., i_{k}, ..., i_{l}, ..., i_{n})(1 < ... < k < ... < l < ... < n)$ be the matrix of $F$ corresponding to the permutation $(i_{1}, ..., i_{k}, ..., i_{l}, ..., i_{n})$ of $A$, then a dual interchange $i_{k} \rightleftharpoons i_{l}$ turns the matrix $M(i_{1}, ..., i_{k}, ..., i_{l}, ..., i_{n})$ into the matrix $M(i_{1}, ..., i_{l}, ..., i_{k}, ..., i_{n})$ of $F$ corresponding to the permutation $(i_{1}, ..., i_{l}, ..., i_{k}, ..., i_{n})$.

\vspace{2mm}
\hspace{-10mm} \textbf{Proof} Let

\[
  M(i_{1}, ..., i_{k}, ..., i_{l}, ..., i_{n}) = \left(\begin{array}{ccccccccccccc}
  a_{1, 1}&.&.&.&a_{1, k}&.&.&.&a_{1, l}&.&.&.&a_{1, n}\\
  .&.&.&.&.&.&.&.&.&.&.&.&\\
  a_{k, 1}&.&.&.&a_{k, k}&.&.&.&a_{k, l}&.&.&.&a_{k, n}\\
  .&.&.&.&.&.&.&.&.&.&.&.&\\
  a_{l, 1}&.&.&.&a_{l, k}&.&.&.&a_{l, l}&.&.&.&a_{l, n}\\
   .&.&.&.&.&.&.&.&.&.&.&.&\\
  a_{n, 1}&.&.&.&a_{n, k}&.&.&.&a_{n, l}&.&.&.&a_{n, n}
  \end{array}\right)
\]

\hspace{-8mm} be the matrix of $F$ corresponding to the permutation $(i_{1}, ..., i_{k}, ..., i_{l}, ..., i_{n})$, where $a_{s, t} = 1$ if and only if $(i_{s}, i_{t}) \in R(1 \leq s, t \leq n)$. If we make a dual interchange $k \rightleftharpoons l$ of the matrix $M(i_{1}, ..., i_{k}, ..., i_{l}, ..., i_{n})$, then the matrix $M(i_{1}, ..., i_{k}, ..., i_{l}, ..., i_{n})$ changes into the following matrix

 \[
  M(i_{1}, ..., i_{k}, ..., i_{l}, ..., i_{n})_{(k \rightleftharpoons l)} = \left(\begin{array}{ccccccccccccc}
  a_{1, 1}&.&.&.&a_{1, l}&.&.&.&a_{1, k}&.&.&.&a_{1, n}\\
  .&.&.&.&.&.&.&.&.&.&.&.&\\
  a_{l, 1}&.&.&.&a_{l, l}&.&.&.&a_{l, k}&.&.&.&a_{l, n}\\
  .&.&.&.&.&.&.&.&.&.&.&.&\\
  a_{k, 1}&.&.&.&a_{k, l}&.&.&.&a_{k, k}&.&.&.&a_{k, n}\\
   .&.&.&.&.&.&.&.&.&.&.&.&\\
  a_{n, 1}&.&.&.&a_{n, l}&.&.&.&a_{n, k}&.&.&.&a_{n, n}
  \end{array}\right)
\]

\vspace{2mm} On the other hand, if we denote the matrix $M(i_{1}, ..., i_{l}, ..., i_{k}, ..., i_{n})$ of $F$ corresponding to the permutation $(i_{1}, ..., i_{l}, ..., i_{k}, ..., i_{n})$ by

\[
  \left(\begin{array}{ccccccccccccc}
  b_{1, 1}&.&.&.&b_{1, k}&.&.&.&b_{1, l}&.&.&.&b_{1, n}\\
  .&.&.&.&.&.&.&.&.&.&.&.&\\
  b_{k, 1}&.&.&.&b_{k, k}&.&.&.&b_{k, l}&.&.&.&b_{k, n}\\
  .&.&.&.&.&.&.&.&.&.&.&.&\\
  b_{l, 1}&.&.&.&b_{l, k}&.&.&.&b_{l, l}&.&.&.&b_{l, n}\\
   .&.&.&.&.&.&.&.&.&.&.&.&\\
  b_{n, 1}&.&.&.&b_{n, k}&.&.&.&b_{n, l}&.&.&.&b_{n, n}
  \end{array}\right),
\]

\vspace{2mm}
\hspace{-8mm} then, by Definition 4, we have $b_{s, t} = 1$ if and only if $(i_{s}, i_{t}) \in R$ when $s \neq k, l$ and $t \neq k, l$; $b_{s, k} = 1$ if and only if $(i_{s}, i_{l}) \in R$, $b_{s, l} = 1$ if and only if $(i_{s}, i_{k}) \in R$, $b_{k, t} = 1$ if and only if $(i_{l}, i_{t}) \in R$, $b_{l, t} = 1$ if and only if $(i_{k}, i_{t}) \in R$. Thus, we have $b_{s, t} = a_{s, t}$ when $s \neq k, l$ and $t \neq k, l$; $b_{s, k} = a_{s, l}$ and $b_{s, l} = a_{s, k}(1 \leq s \leq n)$; $b_{k, t} = a_{l, t}$ and $b_{l, t} = a_{k, t}(1 \leq t \leq n)$. It follows that $M(i_{1}, ..., i_{k}, ..., i_{l}, ..., i_{n})_{(k \rightleftharpoons l)} = M(i_{1}, ..., i_{l}, ..., i_{k}, ..., i_{n})$, and so the proof is done.

\vspace{2mm}
\hspace{-10mm} \textbf{Remark} By the proof of above Lemma, we can see that the dual interchange $k \rightleftharpoons l$ can also turn the matrix $M(i_{1}, ..., i_{l}, ..., i_{k}, ..., i_{n})$ corresponding to the permutation $(i_{1}, ..., i_{l}, ..., i_{k}, ..., i_{n})$ into the matrix $M(i_{1}, ..., i_{k}, ..., i_{l}, ..., i_{n})$ corresponding to the permutation $(i_{1}, ..., i_{k}, ..., i_{l}, ..., i_{n})$. So, for any two matrices corresponding to different permutations of $A$ we can turn one matrix into another by a sequence of dual interchanges.

\vspace{2mm}
\hspace{-10mm} \textbf{Theorem 22} Given an argumentation framework $F = (A, R)$ with $A = \{1, 2, ..., n\}$. Let $M(F)$ be the matrix of $F$ corresponding to the natural permutation $(1, 2, ..., n)$ of $A$, $S = \{i_{1}, i_{2}, ..., i_{k}\} \subset A (1 \leq i_{1} < i_{2} < ... < i_{k} \leq n)$, then by a sequence of dual interchanges we can turn $M(F)$ into the matrix $M(i_{1}, i_{2}, ..., i_{k}, j_{1}, j_{2}, ..., j_{h})$ corresponding to the permutation $(i_{1}, i_{2}, ..., i_{k}, j_{1}, ..., j_{h})$, where $\{j_{1}, j_{2}, ..., j_{h}\} = A \setminus S$ and $1 \leq j_{1} < j_{2} < ... < j_{h} \leq n$.
In particular, the matrix $M(i_{1}, i_{2}, ..., i_{k}, j_{1}, j_{2}, ..., j_{h})$ has the following partition form

\vspace{2mm}\hspace{40mm}
$\left(\begin{array}{ccc}
  M^{cf} & & M^{s} \\
  M^{a} & & M^{c}
 \end{array}\right)$,

\vspace{2mm}
\hspace{-8mm} where $M^{cf}, M^{s}, M^{a}$ and $M^{c}$ are the $cf$-principle sub-block, $s$-sub-block, $M^{a}$-sub-block and $M^{c}$-sub-block of the set $S$ in the matrix $M(F)$, respectively.

\vspace{2mm}
\hspace{-10mm} \textbf{Proof} Let us fist consider the argument $i_{1}$. If $i_{1} = 1$, then no dual interchange is needed, and $M(F) = M(i_{1}, 2, ..., n)$. Otherwise, by making the dual interchange $1 \rightleftharpoons i_{1}$ of the matrix $M(F)$ we get the matrix $M(i_{1}, ..., 1, ...)$ of $F$ corresponding to the permutation $(i_{1}, ..., 1, ...)$, where $1$ is at the position of the natural permutation where $i_{1}$ is at. \\

Secondly, we discuss the argument $i_{2}$. If $i_{1} = 1$ and $i_{2} = 2$, then no dual interchange is needed, and $M(F) = M(i_{1}, i_{2}, 2, ..., n)$. If $i_{1} = 1$ and $i_{2} \neq 2$, by making the dual interchange $2 \rightleftharpoons i_{2}$ we turn the matrix $M(i_{1}, 2, ..., n)$ into the matrix $M(i_{1}, i_{2}, ..., 2, ..., n)$ corresponding to the permutation $i_{1}, i_{2}, ..., 2, ..., n$. If $i_{1} = 2$, then $i_{2} \neq 1$ and the dual interchange $2 \rightleftharpoons i_{2}$ turn the matrix $M(i_{1}, ..., 1, ...) = M(i_{1}, 1, ...) = M(2, 1, ...)$ into the matrix $M(i_{1}, i_{2}, ..., 1, ...) = M(2, i_{2}, ..., 1, ...)$ corresponding to the permutation $(i_{1}, i_{2}, ..., 1, ...)$, where $1$ is at the position of the permutation $(i_{1}, ..., 1, ...) = (i_{1}, 1, ...)$ where $i_{2}$ is at. If $i_{1} \neq 1, 2$, then $i_{2}$ is behind of $1$ in the permutation $(i_{1}, ..., 1, ...)$ and the dual interchange $2 \rightleftharpoons i_{2}$ turn the matrix $M(i_{1}, ..., 1, ...)$ into the matrix $M(i_{1}, i_{2}, ..., 1, ..., 2, ...)$ corresponding to the permutation $(i_{1}, i_{2}, ..., 1, ..., 2, ...)$, where $2$ is at the position of the permutation $(i_{1}, ..., 1, ...)$ where $i_{2}$ is at.

This process can be done step by step. Suppose that we have got the matrix $M(i_{1}, i_{2}, ..., i_{k - 1}, p_{1}, p_{2}, ..., p_{n - k + 1})$ where $(p_{1}, p_{2}, ..., p_{n - k + 1})$ is a permutation of the set $A \setminus \{i_{1}, i_{2}, ..., i_{k - 1}\}$, we finally handle the argument $i_{k}$. If $i_{k} = p_{1}$, then the matrix $M(i_{1}, i_{2}, ..., i_{k - 1}, p_{1}, ..., p_{n - k + 1})$

\hspace{-8mm} $= M(i_{1}, i_{2}, ..., i_{k - 1}, i_{k}, p_{2}, ..., p_{n - k + 1})$. Otherwise, we can make the dual interchange $k \rightleftharpoons i_{k}$ of the matrix $M(i_{1}, i_{2}, ..., i_{k - 1}, p_{1}, p_{2}, ..., p_{n - k + 1})$, and turn the matrix $M(i_{1}, i_{2}, ..., i_{k - 1}, p_{1}, p_{2}, ..., p_{n - k + 1})$ into the matrix

\hspace{-8mm} $M(i_{1}, i_{2}, ..., i_{k}, q_{1}, q_{2}, ..., q_{n - k})$ of $F$ where  $(q_{1}, q_{2}, ..., q_{n - k})$ is a permutation of the set $A \setminus \{i_{1}, i_{2}, ..., i_{k}\}$.

Similar as the above process for $\{i_{1}, i_{2}, ..., i_{k}\}$, we can turn the matrix $M(i_{1}, i_{2}, ..., i_{k}, q_{1}, q_{2}, ..., q_{n - k})$ into the matrix $M(i_{1}, i_{2}, ..., i_{k}, j_{1}, j_{2}, ..., j_{h})$ corresponding to the permutation $(i_{1}, i_{2}, ..., i_{k}, j_{1}, j_{2}, ..., j_{h})$ by a sequence of dual interchanges, where $(j_{1}, j_{2}, ..., j_{h})$ is a permutation of the set $\{q_{1}, q_{2}, ..., q_{n - k}\}$.

Let $M(F) = M(1, 2, ..., n) = (a_{i, j})$, then by Definition 5 we have $a_{i, j} = 1$ iff $(i, j) \in R$. Let $M(i_{1}, i_{2}, ..., i_{k}, j_{1}, j_{2}, ..., j_{h}) = (b_{i, j})$, then $(i_{s}, i_{t}) \in R$ iff $b_{s, t} = 1$ where $1 \leq s, t \leq k$, $(j_{s}, j_{t}) \in R$ iff $b_{k + s, k + t} = 1$ where $1 \leq s, t \leq l$, $(i_{s}, j_{t}) \in R$ iff $b_{s, k + t} = 1$ where $1 \leq s \leq k$ and $1 \leq t \leq l$. It follows that $b_{s, t} = a_{i_{s}, i_{t}}$ where $1 \leq s, t \leq k$, $b_{k+s, k+t} = a_{j_{s}, j_{t}}$ where $1 \leq s, t \leq l$, $b_{s, k+t} = a_{i_{s}, j_{t}}$ where  $1 \leq s \leq k$ and $1 \leq t \leq l$. Therefore, we have

\vspace{2mm}
\hspace{10mm} $M(i_{1}, i_{2}, ..., i_{k}, j_{1}, ..., j_{h}) = \left(\begin{array}{ccc}
  M^{cf} & & M^{s} \\
  M^{a} & & M^{c}
 \end{array}\right)$.

\vspace{2mm}
\hspace{-8mm} where $M^{cf}, M^{s}, M^{a}$ and $M^{c}$ are the $cf$-principle sub-block, $s$-sub-block, $M^{a}$-sub-block and $M^{c}$-sub-block of the set $S$ in the matrix $M(F)$, respectively.

\vspace{2mm}
\hspace{-10mm} \textbf{Corollary 23} Given an argumentation framework $F = (A, R)$ with $A = \{1, 2, ..., n\}$. Let $M(i_{1}, i_{2}, ..., i_{k}, j_{1}, ..., j_{h})$ be the matrix of $F$ correspond to the permutation $(i_{1}, i_{2}, ..., i_{k}, j_{1}, ..., j_{h})$ in Theorem 21, then $S = \{i_{1}, i_{2}, ..., i_{k}\}$ is a conflict-free set iff  $M^{cf}$ is zero. In other words,

\vspace{2mm}
\hspace{10mm} $M(i_{1}, i_{2}, ..., i_{k}, j_{1}, ..., j_{h}) = \left(\begin{array}{ccc}
  0 & & M^{s} \\
  M^{a} & & M^{c}
 \end{array}\right)$.

\vspace{2mm}
\hspace{-10mm} \textbf{Proof} It follows from Theorem 8 and Theorem 22.

\vspace{2mm}
\hspace{-10mm} \textbf{Corollary 24} Given an argumentation framework $F = (A, R)$ with $A = \{1, 2, ..., n\}$. Let $M(i_{1}, i_{2}, ..., i_{k}, j_{1}, ..., j_{h})$ be the matrix of $F$ correspond to the permutation $(i_{1}, i_{2}, ..., i_{k}, j_{1}, ..., j_{h})$ in Theorem 22, then $S = \{i_{1}, i_{2}, ..., i_{k}\}$ is a stable extension iff every column vector of $M^{s}$ is not non-zero.

\vspace{2mm}
\hspace{-10mm} \textbf{Proof} It follows from Theorem 13 and Theorem 22.

\vspace{2mm}
\hspace{-10mm} \textbf{Example 25} Let us continue to consider the argumentation framework $F = (A, R)$ in Example 10. The matrix of $F$ corresponding to the natural permutation $(1,2,3,4,5)$ is

\[
  M(F) = \left(\begin{array}{ccccc}
  0&1&0&0&0\\
  0&0&1&0&1\\
  0&0&0&0&0\\
  0&0&1&0&0\\
  0&0&0&1&0
  \end{array}\right).
\]

As have been shown in $Ex.12$, $\mathcal{S}(0) = \{\emptyset \}$, $\mathcal{S}(1) = \{ \{1\}, \{2\}, ..., \{5\} \}$,

\hspace{-8mm} $\mathcal{S}(2) = \{\{1, 3\}, \{1, 4\},\{1, 5\}, \{2, 4\}, \{3, 5\}\}$, $\mathcal{S}(3) = \{1, 3, 5\}$,
$\mathcal{S}(4)$ and $\mathcal{S}(5)$.

For $S = \{1,3\}$, by taking the interchange $2 \rightleftharpoons 3$ the matrix $M(F)$ is turned into the matrix

\[
  M(1,3,2,4,5) = \left(\begin{array}{ccccc}
  0&0&1&0&0\\
  0&0&0&0&0\\
  0&1&0&0&1\\
  0&1&0&0&0\\
  0&0&0&1&0
  \end{array}\right),
\]
where
\[
  M^{cf} = \left(\begin{array}{cc}
  0&0\\
  0&0
  \end{array}\right), M^{s} = \left(\begin{array}{ccc}
  1&0&0\\
  0&0&0
  \end{array}\right), M^{a} = \left(\begin{array}{ccc}
  0&1\\
  0&1\\
  0&0
  \end{array}\right), M^{c} = \left(\begin{array}{ccc}
  0&0&1\\
  0&0&0\\
  0&1&0
  \end{array}\right).
\]
Since not every column vector of $M^{s}$ is zero, the set $S$ is not a stable extension. Similar discussion show that other elements of $\mathcal{S}(2)$ are also not stable extensions.

For $S^{\prime} = \{1,3,5\}$, by taking the interchange $3 \rightleftharpoons 5$ the matrix $M(1,3,2,4,5)$ is turned into the matrix

\[
  M(1,3,5,4,2) = \left(\begin{array}{ccccc}
  0&0&0&0&1\\
  0&0&0&0&0\\
  0&0&0&1&0\\
  0&1&0&0&0\\
  0&1&1&0&0
  \end{array}\right).
\]
where
\[
  M^{cf} = \left(\begin{array}{ccc}
  0&0&0\\
  0&0&0\\
  0&0&0
  \end{array}\right), M^{s} = \left(\begin{array}{ccc}
  0&1\\
  0&0\\
  1&0
  \end{array}\right), M^{a} = \left(\begin{array}{ccc}
  0&1&0\\
  0&1&1
  \end{array}\right), M^{c} = \left(\begin{array}{cc}
  0&0\\
  0&0
  \end{array}\right).
\]
In this case, the column vectors of $M^{s}$ are all non-zero. And thus, $S^{\prime}$ is a stable extension.

\vspace{2mm}
\hspace{-10mm} \textbf{Theorem 26} Given an argumentation framework $F = (A, R)$ with $A = \{1, 2, ..., n\}$. Let $M(F)$ be the matrix of $F$ corresponding to the natural permutation $(1, 2, ..., n)$, then $S = \{i_{1}, i_{2}, ..., i_{k}\}$ is a conflict-free set iff by a sequence of dual interchanges we can turn $M(F)$ into the following partition matrix:

\vspace{2mm}
\hspace{30mm} $\left(\begin{array}{ccc}
  O_{k,k} & O_{k,q} & S_{k,l} \\
  A_{q,k} & C_{q,q} & E_{q,l} \\
  F_{l,k} & G_{l,q} & H_{l,l}
 \end{array}\right) $

\vspace{2mm}
\hspace{-8mm} where each column vector of $S_{k, l}$ is non-zero, $k + q + l = n$ and $q \geq 0$.

\vspace{2mm}
\hspace{-10mm} \textbf{Proof} From Corollary 23, $S = \{i_{1}, i_{2}, ..., i_{k}\}$ is a conflict-free set iff by a sequence of dual interchanges $M(F)$ can be turned into the partition matrix

\vspace{2mm}
\hspace{10mm} $M(i_{1}, i_{2}, ..., i_{k}, j_{1}, ..., j_{h}) = \left(\begin{array}{ccc}
  0 & & M^{s} \\
  M^{a} & & M^{c}
 \end{array}\right)$

\vspace{2mm}
\hspace{-8mm} corresponding to the permutation $(i_{1}, i_{2}, ..., i_{k}, j_{1}, ..., j_{h})$, where $M^{s}, M^{a}$ and $M^{c}$ are the $s$-sub-block, $M^{a}$-sub-block and $M^{c}$-sub-block of the set $S$ in the matrix $M(F)$, respectively.

If there is no zero column in the sub-block $M^{s}$ of $M(i_{1}, i_{2}, ..., i_{k}, j_{1}, ..., j_{h})$, then $q = 0$ and we have done. Otherwise, we may assume that all the zero column vectors of $M^{s}$ are the columns $t_{1}, t_{2}, ..., t_{q}(1 \leq q \leq h)$. Certainly, they correspond to the columns $k + t_{1}$, $k + t_{2}$, ..., $k + t_{q}$ of the matrix $M(i_{1}, i_{2}, ..., i_{k}, j_{1}, ..., j_{h})$ respectively.

Similar as the proof of Theorem 22, after making a sequence of dual interchanges the matrix $M(i_{1}, i_{2}, ..., i_{k}, j_{1}, ..., j_{h})$ shall be turned into the matrix $M(i_{1}, i_{2}, ..., i_{k}, j_{t_{1}}, ..., j_{t_{q}}, j_{s_{1}}, ..., j_{s_{l}})$ corresponding to the permutation $(i_{1}, i_{2}, ..., i_{k}, j_{t_{1}}, ..., j_{t_{q}}, j_{s_{1}}, ..., j_{s_{l}})$, where $(j_{s_{1}}, ..., j_{s_{l}})$ is a permutation of the set $A \setminus \{i_{1}, i_{2}, ..., i_{k}, j_{t_{1}}, ..., j_{t_{q}}\}$. By now, the elements at the intersection of first $k$ rows and first $k+q$ columns of $M(i_{1}, i_{2}, ..., i_{k}, j_{t_{1}}, ..., j_{t_{q}}, j_{s_{1}}, ..., j_{s_{l}})$ are zero. Therefore, the matrix $M(i_{1}, i_{2}, ..., i_{k}, j_{t_{1}}, ..., j_{t_{q}}, j_{s_{1}}, ..., j_{s_{l}})$ has the following partition form:

\vspace{2mm}
\hspace{30mm} $\left(\begin{array}{ccc}
  O_{k,k} & O_{k,q} & S_{k,l} \\
  A_{q,k} & C_{q,q} & E_{q,l} \\
  F_{l,k} & G_{l,q} & H_{l,l}
 \end{array}\right) $

\vspace{2mm}
\hspace{-8mm} where each column vector of $S_{k, l}$ is non-zero, $k + q + l = n$ and $q \geq 1$.

\vspace{2mm}
\hspace{-10mm} \textbf{Remark} Since the partition matrices obtained in Corollary 23 and Theorem 26 play a central role in finding out the extensions of $F = (A, R)$, we called them the norm form of the matrix $M(F)$ of $F = (A, R)$.

\vspace{2mm}
\hspace{-10mm} \textbf{Corollary 27} Given an argumentation framework $F = (A, R)$ with $A = \{1, 2, ..., n\}$. Let $M(i_{1}, i_{2}, ..., i_{k}, j_{t_{1}}, ..., j_{t_{q}}, j_{s_{1}}, ..., j_{s_{l}}) = (b_{i, j})$ be the norm form of $M(F)$ corresponding to the permutation $(i_{1}, i_{2}, ..., i_{k}, j_{t_{1}}, ..., j_{t_{q}}, j_{s_{1}}, ..., j_{s_{l}})$ in Theorem 26, then, $S = \{i_{1}, i_{2}, ..., i_{k}\}$ is an admissible extension iff $A_{q,k} = 0$.  \\

\vspace{2mm}
\hspace{-10mm} \textbf{Proof} Necessity. Assume that $A_{q,k} \neq 0$, then there are some $r(1 \leq r \leq k)$ and $v(1 \leq u \leq q)$ such that the element at the intersection of the $u$-th row and the $r$-th column in $A_{q,k}$ is $1$, which is at the intersection of the $(k + u)$-th row and the $r$-th column in $M(i_{1}, i_{2}, ..., i_{k}, j_{t_{1}}, ..., j_{t_{q}}, j_{s_{1}}, ..., j_{s_{l}})$. It follows that $b_{k+u, r} = 1$, i.e., $(j_{t_{u}}, i_{r}) \in R$. Thus, the argument $i_{r}$ is attacked by $j_{s_{u}}$. But, from the zero sub-block $O_{k, q}$ we know that $b_{w, k+u} = 0$ for any $1 \leq w \leq k$. So, there is no argument $i_{w}(1 \leq w \leq k)$ in $S$ which attacks $j_{t_{u}}$. And thus, the argument $i_{r}$ is not defended by $S$, a contradiction with the hypothesis

Sufficiency. Obviously, $S$ is a conflict-free set in terms of the zero sub-block $O_{k,k}$ in $M(i_{1}, i_{2}, ..., i_{k}, j_{t_{1}}, ..., j_{t_{q}}, j_{s_{1}}, ..., j_{s_{l}})$. Since $A_{q,k} = 0$, we have $b_{k+u, r} = 0(1 \leq r \leq k, 1 \leq u \leq q)$, i.e., $(j_{t_{u}}, i_{r}) \notin R$. Thus, $S$ is not attacked by any argument of the set $\{j_{t_{1}}, ..., j_{t_{q}}\}$.

Let $i_{r}(1 \leq r \leq k)$ be any fixed argument of $S$, if there is some argument $p$ which attacks $i_{r}$, then $p =  j_{s_{v}}$ for some $(1 \leq v \leq l)$. Since the $v$-th column vector in $S_{k,l}$ is not zero, there is some $w(1 \leq w \leq k)$ such that $b_{w, k+q+v} = 1$. It follows that $(i_{w}, j_{s_{v}}) \in R$, i,e., $p = j_{s_{v}}$ is attacked by the argument $i_{w}$ of $S$. Therefore, the conflict-free set $S$ is defended by itself, and thus an admissible extension.

\vspace{2mm}
\hspace{-10mm} \textbf{Corollary 28} Given an argumentation framework $F = (A, R)$ with $A = \{1, 2, ..., n\}$. Let $M(i_{1}, i_{2}, ..., i_{k}, j_{t_{1}}, ..., j_{t_{q}}, j_{s_{1}}, ..., j_{s_{l}})$ be the norm form of $M(F)$ corresponding to the permutation $(i_{1}, i_{2}, ..., i_{k}, j_{t_{1}}, ..., j_{t_{q}}, j_{s_{1}}, ..., j_{s_{l}})$ in Theorem 26, then $S = \{i_{1}, i_{2}, ..., i_{k}\}$ is a complete extension iff $A_{q,k} = 0$ and each column vector of $C_{q,q}$ is not zero.

\vspace{2mm}
\hspace{-10mm} \textbf{Proof} Sufficiency. First, $S$ is an admissible extension follows from $A_{q,k} = 0$. Since each column vector of $S_{k, l}$ is non-zero, for every $v(1 \leq v \leq l)$ there is some $r(1 \leq r \leq k)$ such that $b_{r, k+q+v} = 1$, i.e., $(i_{r}, j_{s_{v}}) \in R$. Thus, the argument $j_{s_{v}}(1 \leq v \leq l)$ is not defended by $S$.

Let us consider any fixed argument $j_{t_{u}}(1 \leq u \leq q)$. Because each column vector of $C_{q,q}$ is not zero, there is some $v(1 \leq v \leq q)$ such that $b_{k+v, k+u} = 1$, i.e., $(j_{t_{v}}, j_{t_{u}}) \in R$. On the other hand, $b_{r, k+v} = $ for every $1 \leq r \leq k$ because of $O_{k, q}$. So, $(i_{r}, j_{t_{v}}) \notin R(1 \leq r \leq k)$. That means the argument $j_{t_{v}}$ is not attacked by any element of $S$, and thus we conclude that $j_{t_{u}}$ is not defended by $S$.

To sum up, $S$ is a complete extension of $F$.\\

Necessity. Assume that $S$ is a complete extension of $F$, then $S$ is a conflict-free set and thus by a sequence of dual interchanges the matrix $M(F)$ can be turn into the following the norm form

\vspace{2mm}
\hspace{30mm} $\left(\begin{array}{ccc}
  O_{k,k} & O_{k,q} & S_{k,l} \\
  A_{q,k} & C_{q,q} & E_{q,l} \\
  F_{l,k} & G_{l,q} & H_{l,l}
 \end{array}\right) $

\vspace{2mm}
\hspace{-8mm} where each column vector of $S_{k, l}$ is non-zero, $k + q + l = n$ and $q \geq 1$. Obviously, $A_{q,k} = 0$ comes from the fact that $S$ is an admissible extension of $F$. Next, we shall prove that each column vector of $C_{q, q}$ is not zero.

If there is some $u(1 \leq u \leq q)$ such that the $u$-th column vector of $C_{q, q}$ is zero, then $b_{k+v, k+u} = 0$ for each $v(1 \leq v \leq q)$, i.e., $(j_{t_{v}}, j_{t_{u}}) \notin R(1 \leq v \leq q)$. Considering the sub-block $O_{k,q}$, we also have $b_{r, k+u} = 0$ for each $r(1 \leq r \leq k)$, i.e., $(i_{r}, j_{t_{u}}) \notin R(1 \leq r \leq k)$. Let $p$ be an attacker of the argument $j_{t_{u}}$, then there must be  some $w(1 \leq w \leq l)$ such that $p = j_{s_{w}}$. But, we know that each column vector of $S_{k,l}$ is non-zero, and thus the $w$-th column vector of $S_{k,l}$ is non-zero. So, there is some $r(1 \leq r \leq k)$ such that $b_{r, k+q+w} = 1$, i.e., $(i_{r}, j_{s_{w}}) \in R$. That means the argument $i_{r}$ is an attacker of $p = j_{s_{w}}$. Therefore, the argument $j_{t_{u}}$ is defended by $S$, which contradicts with the fact that $S$ is a complete extension.

\vspace{2mm}
\hspace{-10mm} \textbf{Example 29} Consider the argumentation framework $F = (A, R)$, where $A = \{1,2,3,4,5\}$ and $R = \{(1,2),(1,3),(3,1),(4,5),(5,1),(5,4)\}$. Then, the matrix of $F$ corresponding to the natural permutation $(1,2,3,4,5)$ is

\[
  M(F) = \left(\begin{array}{ccccc}
  0&1&1&0&0\\
  0&0&0&0&0\\
  1&0&0&0&0\\
  0&0&0&0&1\\
  1&0&0&1&0
  \end{array}\right).
\]

By Theorem 9, it is easy to fin out all the conflict-free sets: $\{1,4\}$, $\{1,5\}$, $\{2,3\}$£¬$\{2,4\}$£¬$\{2, 5\}$£¬$\{3,4\}$£¬$\{3,5\}$£¬ $\{1,4,2\}$£¬$\{1,4,5\}$£¬$\{2,3,4\}$£¬$\{2,3,5\}$.

\vspace{2mm} For the set $S = \{3, 4\}$, we make the dual interchange $1 \rightleftharpoons 3$ and turn the matrix $M(F)$ into the matrix

\[
  M(3,2,1,4,5) = \left(\begin{array}{ccccc}
  0&0&1&0&0\\
  0&0&0&0&0\\
  1&1&0&0&0\\
  0&0&0&0&1\\
  0&0&1&1&0
  \end{array}\right).
\]
Furthermore, by making the dual interchange $2 \rightleftharpoons 4$ on the matrix $M(3,2,1,4,5)$ we turn the matrix $M(3,2,1,4,5)$ into the matrix
\[
  M(3,4,1,2,5) = \left(\begin{array}{ccccc}
  0&0&1&0&0\\
  0&0&0&0&1\\
  1&0&0&1&0\\
  0&0&0&0&0\\
  0&1&1&0&0
  \end{array}\right),
\]
where
\[
  M^{s} = \left(\begin{array}{ccc}
  1&0&0\\
  0&0&1
  \end{array}\right), M^{a} = \left(\begin{array}{ccc}
  1&0\\
  0&0\\
  0&1
  \end{array}\right), M^{c} = \left(\begin{array}{ccc}
  0&1&0\\
  0&0&0\\
  1&0&0
  \end{array}\right).
\]
In this case, the column vectors of $M^{s}$ corresponding to the non-zero row vectors of $M^{a}$ are non-zero. Thus, $S$ is an admissible extension according to Theorem 16.

Next, by making the dual interchange $3 \rightleftharpoons 4$ on the matrix $M(3,4,1,2,5)$ we turn the matrix $M(3,2,1,4,5)$ into the matrix
\[
  M(3,4,2,1,5) = \left(\begin{array}{ccccc}
  0&0&0&1&0\\
  0&0&0&0&1\\
  0&0&0&0&0\\
  1&0&1&0&0\\
  0&1&0&1&0
  \end{array}\right),
\]
where
\[
  A_{1,2} = \left(\begin{array}{cc}
  0&0
  \end{array}\right), S_{2,2} = \left(\begin{array}{cc}
  1&0\\
  0&1
  \end{array}\right), C_{1,1} = \left(\begin{array}{cc}
  0
  \end{array}\right), E_{1,2} = \left(\begin{array}{cc}
  0&0
  \end{array}\right), F_{2,2} = \left(\begin{array}{cc}
  1&0\\
  0&1
  \end{array}\right),
\]

\[ \hspace{-80mm} G_{2,1} = \left(\begin{array}{cc}
  1\\
  0
  \end{array}\right), H_{2,2} = \left(\begin{array}{cc}
  0&0\\
  1&0
  \end{array}\right).
\]
In this case, $A_{2,1} = 0$ and the only column vector of $C_{1,1}$ is zero. Thus, $S$ is not a complete extension according to Corollary 28.

\vspace{2mm} For the set $S' = \{2,3\}$, we make the dual interchange $1 \rightleftharpoons 3$ and turn the matrix $M(F)$ into the matrix

\[
  M(3,2,1,4,5) = \left(\begin{array}{ccccc}
  0&0&1&0&0\\
  0&0&0&0&0\\
  1&1&0&0&0\\
  0&0&0&0&1\\
  0&0&1&1&0
  \end{array}\right).
\]
where
\[
  M^{s} = \left(\begin{array}{ccc}
  1&0&0\\
  0&0&0
  \end{array}\right), M^{a} = \left(\begin{array}{ccc}
  1&1\\
  0&0\\
  0&0
  \end{array}\right), M^{c} = \left(\begin{array}{ccc}
  0&0&0\\
  0&0&1\\
  1&1&0
  \end{array}\right).
\]
In this case, the column vectors of $M^{s}$ corresponding to the non-zero row vectors of $M^{a}$ are non-zero. Thus, $S$ is an admissible extension according to Theorem 16.

Furthermore, by making the dual interchange $3 \rightleftharpoons 5$ the matrix $M(3,2,1,4,5)$ is turned into the matrix
\[
  M(3,2,5,4,1) = \left(\begin{array}{ccccc}
  0&0&0&0&1\\
  0&0&0&0&0\\
  0&0&0&1&1\\
  0&0&1&0&0\\
  1&1&0&0&0
  \end{array}\right),
\]
where
\[
  A_{2,2} = \left(\begin{array}{cc}
  0&0\\
  0&0
  \end{array}\right), S_{2,1} = \left(\begin{array}{cc}
  1\\
  0
  \end{array}\right), C_{2,2} = \left(\begin{array}{cc}
  0&1\\
  1&0
  \end{array}\right), E_{2,1} = \left(\begin{array}{cc}
  1\\
  0
  \end{array}\right), F_{1,2} = \left(\begin{array}{cc}
  1&1
  \end{array}\right),
\]

\[ \hspace{-80mm} G_{1,2} = \left(\begin{array}{cc}
  1&0
  \end{array}\right), H_{1,1} = \left(\begin{array}{cc}
  0
  \end{array}\right).
\]
In this case, $A_{2,2} = 0$ and the column vectors of $C_{2,2}$ are all non-zero. Thus, $S$ is a complete extension.  \\

\hspace{-10mm} {\bf \Large 9. Further discussion and related work}

\vspace{5mm} In the above Sections, we have established a matrix approach to find out all the stable(admissible, complete) extensions of an AF. About other common semantics of an AF not mentioned(such as preferred, grounded, ideal, semi-stable and eager extension), we can also find out all the extensions through the matrix approach and additional work which only concerns the comparison of different sets. The procedure consists of two steps: Finding out all the extensions(admissible or complete according to the need) by the matrix approach; Comparing the related sets which are used to define the semantics, and finding out the needed extensions in light of the definition of related extensions.
For example, if we want to find out all the preferred extensions of an argumentation framework $F = (A, R)$, the procedure is as follows. Firstly, finding out all the complete extensions of $F = (A, R)$ by the matrix approach. Then, comparing all the complete extensions from the view of sets,  all the maximal sets are exactly the total preferred extensions we look for.

Now, all the "global" questions concerning Dung's argumentation frameworks proposed by Modgil and Caminada[15] can be solved through the matrix approach. For the "local" questions, the matrix approach is still valid. We only need to make some comparing between the related sets after finding out all the extensions under a given semantics. If we want to decide whether a set $A \in \mathcal{A}$ is contained in an specific extension, we only need to compare the set $A$ with this extension, which has been found out by the matrix approach. For other "local" questions, we can give the similar process based on each specific question.

\vspace{2mm} There are several attempts in the literature for computing extensions of an argumentation framework. Modgil and Caminada have developed the graph labelling approach which was originally proposed by Pollok[17]. Argument game approach is another efficient tool which is based on the proof theories. The constraint satisfaction approach was built by Amgoud and Devred[1]. But, to our knowledge no attempt was done in using the matrix to computer the extensions of AFs.

In [15], the authors summarised the labelling approach by which the core semantics of AFs defined by Dung and others[9, 5, 4, 10] can be found out. But, this approach has an obvious drawback: The admissible extension found out by graph labelling approach depend on the selection of the elements that are illegally IN, so different selections may lead to same admissible extension. In particular, there is no way to know whether the admissible extensions have been found out entirely. The argument game approach mainly focus on solving the "local" questions, but only a selection of the "local" questions have been answered just as Modgil and Caminada described in [15]. The constraint satisfaction approach possess more technical feature. It encodes an AF as a Constraint Satisfaction Problem, and thus is able to use some powerful solvers for computing the extensions of the argumentation framework. The problem lies in how to find the candidate extensions which we want to verify by the criterion established in their paper.

\vspace{2mm} Our matrix approach first find out all the conflict-free sets of an AF, then turn the matrix of the AF into a norm form with respect to a specific semantics(stable, admissible or complete), finally select out all the extensions according to the related criterions corresponding to different semantics. For other semantics, such as grounded extension, preferred extension, ideal extension, semi-stable extension and eager extension, we can find out them from the related family of extensions by verifying whether they satisfy the conditions(only concerning the comparison of different sets) in their definitions. With regard to the "local" questions, the remaining work is only to compare the inclusion relations of deferent sets of each question, after finding out all the needed extensions.  \\

\hspace{-10mm} {\bf \Large 10. Conclusions and perspectives}

\vspace{5mm} In this paper, we introduce the matrix representation $M(F)$ of an argumentation framework $F = (A, R)$, and the $cf$-block $M^{cf}$, $s$-block $M^{s}$, $a$-block $M^{a}$ and $c$-block $M^{c}$ with respect to a set $S \subset A$. Several several theorems have been established in order to determine the core extensions (stable, admissible, complete) of an argumentation framework, by sub-blocks of the matrix $M(F)$ and the relations between these sub-blocks. Furthermore, we propose a matrix approach finding out all the extensions of an argumentation framework under a given semantics (stable, admissible, complete). For other semantics, we can also compute all the related extensions if we combine the matrix approach with some additional work concerning the comparison of different sets.

Interestingly, the $s$-block $M^{s}$ ($a$-block $M^{a}$, $c$-block $M^{c}$) of the set $S$ correspond to the determination for $S$ to be a stable extension (admissible extension, complete extension respectively). And, the $c$-block of $S$ is exactly the complementary sub-block of the $cf$-block of $S$, the $a$-block of $S$ is exactly the complementary sub-block of the $s$-sub-block of $S$. In addition, the dual interchanges provide us a chance to turn the matrix $M(F)$ of the argumentation framework $F = (A, R)$ into a norm form which can be easily employed to determine whether a conflict-free set is a specific extension.

The prospectives are that, we can introduce or build more matrix tools into the research of argumentation frameworks. Our future goal is to develop the matrix approach in the related areas, such as bipolar argumentation frameworks, fuzzy argumentation frameworks. Anther direction is to set up the bridge from argument game to matrix operations, so as to find the possibility of expressing the argument game by matrices[13, 14, 19].   \\


\hspace{-8mm}{\large \bf  References}
\hspace{5mm}



\begin{enumerate}

\bibitem{s1} L. Amgoud, C. Devred, Argumentation frameworks as Constraint Satisfaction Problems, In \emph{Proc. SUM}, volume 6929 of \emph{LNCS}, 2011, 110-122. Springer.

\bibitem{s2} P. Baroni, M. Giacomin, On principle-based evaluation of extension-based argumentation semantics, Artificial Intelligence 171 (2007), 675-700.

\bibitem{s3} T. J. M. Bench-Capon, Paul E. Dunne, Argumentation in artificial intelligence, Artificial intelligence 171(2007)619-641

\bibitem{s4} M. Caminada, Semi-stable semantics, in: Frontiers in Artificial Intelligence and its Applications, vol. 144, IOS Press, 2006, pp. 121-130.

\bibitem{s5} C. Cayrol, M. C. Lagasquie-Schiex, Graduality in argumentation, J. AI Res. 23 (2005)245-297.

\bibitem{s6} S. Coste-Marquis, C. Devred, Symmetric argumentation frameworks, in: Lecture Notes in Artificial Intelligence, vol. 3571, Springer-Verlag, 2005, pp. 317-328.

\bibitem{s7} S. Coste-Marquis, C.Devred, P. Marquis, Prudent semantics for argumentation frameworks, in: Proc. 17th ICTAI, 2005, pp. 568-572.

\bibitem{s8} Y. Dimopoulos, A. Torres, Graph theoretical structures in logic programs and default theories, Teoret. Comput. Sci. 170(1996)209-244.

\bibitem{s9} P. M. Dung, On the acceptability of arguments and its fundamental role in nonmonotonic reasoning, logic programming and $n$-person games, Artificial Intelligence 77 (1995), 321-357.

\bibitem{s10} P. M. Dung, P. Mancarella, F. Toni, A dialectic procedure for sceptical assumption-based argumentation, in: Frontiers in Artificial Intelligence and its Applications, vol. 144, IOS Press, 2006, pp. 145-156.

\bibitem{s11} P. E. Dunne, Computational properties of argument systems satisfying graph-theoretic constrains, Artificial Intelligence 171 (2007), 701-729.

\bibitem{s12} P. E. Dunne, T. J. M. Bench-Capon, Coherence in finite argument systems, Artificial intelligence 141(2002)187-203.


\bibitem{s13} P. E. Dunne, T. J. M. Bench-Capon, Two party immediate response disputes: properties and efficiency, Artificial Intelligence 149 (2003), 221-250.

\bibitem{s14}  H. Jakobovits, D. Vermeir, Dialectic semantics for argumentation frameworks, in: Proc. 7th ICAIL, 1999, pp. 53-62.

\bibitem{s15}  S. Modgil, M. Caminada, Proof Theories and Algorithms for Abstract Argumentation Frameworks,  In: Rahwan I., Simari G, editors. Argumentation in AI. Springer; 2009. p. 105-129.


\bibitem{s16} E. Oikarinen, S. Woltron, Characterizing strong equivalence for argumentation frameworks, Artificial intelligence(2011), doi:10.1016/j.artint. 2011.06.003.

\bibitem{s17}  J. L. Pollock, Cognitive Carpentry, A Blueprint for How to Build a Person, MIT Press, Cambridge, MA, 1995.


\bibitem{s18} G. Vreeswijk, Abstract argumentation system, Artificial intelligence 90(1997)225-279.

\bibitem{s19} G. Vreeswijk, H. Pakken, Credulous and sceptical argument games for preferred semantics, in: Proceedings of JELIA'2000, the 7th European Workshop on Logic for Artificial Intelligence, Berlin, 2000, pp. 224-238.




\end{enumerate}
\end{document}